\documentclass[10pt,twocolumn,letterpaper]{article}

\usepackage{cvpr}              % To produce the CAMERA-READY version
\usepackage[dvipsnames]{xcolor}

\definecolor{cvprblue}{rgb}{0.21,0.49,0.74}
\usepackage[pagebackref,breaklinks,colorlinks,allcolors=magenta]{hyperref}
\usepackage{amssymb,bbding,pifont}

\definecolor{myblue}{RGB}{66,133,244}
\definecolor{mygreen}{RGB}{51,168,83}
\definecolor{myyellow}{RGB}{251,188,3}
\definecolor{myred}{RGB}{234,67,53}
\definecolor{mygrey}{RGB}{95,99,104}
\definecolor{mypup}{RGB}{153,0,204}

\def\paperID{*****} % *** Enter the Paper ID here
\def\confName{CVPR}
\def\confYear{2025}
\hypersetup{breaklinks,colorlinks}
\title{NTIRE 2026 The Second Challenge on Day and Night Raindrop Removal for Dual-Focused Images: Methods and Results}

\author{Xin Li$^{\dagger}$ \quad Yeying Jin$^{\dagger}$ \quad Suhang Yao$^{\dagger}$ \quad Beibei Lin$^{\dagger}$ \quad Zhaoxin Fan$^{\dagger}$ \quad Wending Yan$^{\dagger}$ \quad Xin Jin$^{\dagger}$ \\ Zongwei Wu$^{\dagger}$ \quad Bingchen Li$^{\dagger}$ \quad Peishu Shi$^{\dagger}$ \quad Yufei Wang$^{\dagger}$ \\ Yu Li$^{\dagger}$ \quad Zhibo Chen$^{\dagger}$ \quad Bihan Wen$^{\dagger}$ \quad Robby T. Tan$^{\dagger}$ \quad Radu Timofte$^{\dagger}$ \\
Runzhe Li \quad Kui Jiang \quad Zhaocheng Yu \quad Yiang Chen \quad Junjun Jiang \quad Xianming Liu \\
Hongde Gu \quad Zeliang Li \quad Mache You \quad Jiangxin Dong \quad Jinshan Pan \\
Qiyu Rong \quad Bowen Shao \quad Hongyuan Jing \quad Mengmeng Zhang \quad Bo Ding \quad Hui Zhang \quad Yi Ren \\
Mohab Kishawy \quad Jun Chen \\
Anh\textendash Kiet Duong \quad Petra Gomez\textendash Kr\"amer \quad Jean\textendash Michel Carozza \\
Wangzhi Xing \quad Xin Lu \quad Enxuan Gu \quad Jingxi Zhang \quad Diqi Chen \quad Qiaosi Yi \quad Bingcai Wei \\
Wenjie Li \quad Bowen Tie \quad Heng Guo \quad Zhanyu Ma \\
Jiachen Tu \quad Guoyi Xu \quad Yaoxin Jiang \quad Cici Liu \quad Yaokun Shi \\
Paula Garrido Mellado \quad Daniel Feijoo \quad Alvaro García Lara \quad Marcos V. Conde \\
Zhidong Zhu \quad Bangshu Xiong \quad Qiaofeng Ou \quad Zhibo Rao \quad Wei Li \\
Zida Zhang \quad Hui Geng \quad Qisheng Xu \quad Xuyao Deng \quad Changjian Wang \quad Kele Xu \\
Guanglu Dong \quad Qiyao Zhao \quad Tianheng Zheng \quad Chunlei Li \quad Lichao Mou \quad Chao Ren \\
Chang\textendash De Peng \quad Chieh\textendash Yu Tsai \quad Guan\textendash Cheng Liu \quad Li\textendash Wei Kang \\
Abhishek Rajak \quad Milan Kumar Singh \quad Ankit Kumar \quad Dimple Sonone \quad Kishor Upla \quad Kiran Raja \\
Huilin Zhao \quad Xing Xu \quad Chuan Chen \quad Yeming Lao \quad Wenjing Xun \quad Li Yang \\
Bilel Benjdira \quad Anas M.\ Ali \quad Wadii Boulila \\
Hao Yang \quad Ruikun Zhang \quad Liyuan Pan
}

\begin{document}
\maketitle
\renewcommand{\thefootnote}{}
\footnotetext{$^{\dagger}$X. Li, Y. Jin, S. Yao, B. Lin, Z. Fan, W. Yan, X. Jin, Z. Wu, B. Li, P. Shi, Y. Wang, Y. Li, Z. Chen, B. Wen, R. Tan and R. Timofte are the challenge organizers. The contact details can be found in section~\ref{organizers}.}
\footnotetext{The other authors are participants of the NTIRE 2026 The Second Challenge on Day and Night Raindrop Removal for Dual-Focused
Images.}
\footnotetext{The NTIRE2026 website:~\url{https://cvlai.net/ntire/2026/}}
\footnotetext{The Competition website~\url{https://www.codabench.org/competitions/12808/}}
\footnotetext{The Raindrop Clarity database:~\url{https://github.com/jinyeying/RaindropClarity}}

\begin{abstract}
This paper presents an overview of the NTIRE 2026 Second Challenge on Day and Night Raindrop Removal for Dual-Focused Images. Building upon the success of the first edition, this challenge attracted a wide range of impressive solutions, all developed and evaluated on our real-world Raindrop Clarity dataset~\cite{jin2024raindrop}. For this edition, we adjust the dataset with 14,139 images for training, 407 images for validation, and 593 images for testing. The primary goal of this challenge is to establish a strong and practical benchmark for the removal of raindrops under various illumination and focus conditions. In total, 168 teams have registered for the competition, and 17 teams submitted valid final solutions and fact sheets for the testing phase. The submitted methods achieved strong performance on the Raindrop Clarity dataset, demonstrating the growing progress in this challenging task. The project website is available at~\url{https://lixinustc.github.io/CVPR-NTIRE2026-RainDrop-Competition.github.io/}.
\end{abstract}    
% \section{Introduction}
\label{sec:intro}
% Image deraining has attracted lots of 
\section{Introduction}

Image deraining is a fundamental problem in low-level vision, aiming to restore clear visual content from images degraded by rain streaks~\cite{yang2017deep-rain100} or adherent raindrops~\cite{qian2018attentiveRainDrop}. Beyond improving visual quality, it also benefits downstream vision applications such as autonomous driving, surveillance, and pedestrian analysis. With the rapid development of restoration backbones, including CNNs, Transformers, MLPs, and more recently Mamba- and diffusion-based frameworks~\cite{wang2019spatialSPAData,zamir2022restormer_deblur-transformer,tu2022maxim,wu2024rainmamba,valanarasu2022transweather,li2023learningDIL,pang2020fan,li2020learningCNNderain,zou2024freqmambaDerain,liang2021swinir,li2022hst,ozdenizci2023restoringweatherdiff,shen2023rethinkingRainDiff,chen2024teachingt3diffusion}, image deraining has evolved from conventional single-task restoration to more challenging settings that require stronger robustness, generalization, and perceptual quality. In particular, recent studies have started to explore unified restoration under multiple adverse weather conditions and perceptually oriented deraining methods, further expanding the scope and difficulty of this research area~\cite{valanarasu2022transweather,ozdenizci2023restoringweatherdiff,yu2024sfiqa,zhang2018lpips,lu2024aigcvqa}.

Despite these advances, the progress of raindrop removal in real-world scenarios is still heavily constrained by the availability of suitable datasets. Early studies mainly relied on synthetic data~\cite{hao2019learningRaindropdataset,li2016rainRain12,yang2017deepRain100HL,fu2017removingDDN-data,zhang2018densityDID-Data,li2019singleRain800,hu2019depthRainCityScapes,li2019heavyOutdoor-Rain,jiang2020multiRain13K}, since it is difficult to capture paired rainy and clean images under controlled real conditions. More recent efforts have attempted to build realistic datasets by exploiting temporal priors in videos~\cite{wang2019spatialSPAData} or by physically simulating raindrops on transparent media~\cite{qian2018attentiveRaindropdataset,soboleva2021raindropsRaindropdataset,porav2019canRaindropdataset,quan2021removingRainDS}. However, existing benchmarks still have notable limitations. Most of them focus on daytime scenes, place limited emphasis on adherent raindrops in complex real environments, and rarely consider different focusing modes. In particular, datasets containing both raindrop-focused and background-focused images, especially under both daytime and nighttime conditions, remain scarce~\cite{lin2024nightrain,jin2022unsupervised,jin2023enhancing,lin2024nighthaze}.

To facilitate research in this direction, following the first edition, we organize the NTIRE 2026 Second Challenge on Day and Night Raindrop Removal for Dual-Focused Images. The challenge is built upon the Raindrop Clarity dataset~\cite{jin2024raindrop}, a real-world benchmark designed to cover both daytime and nighttime scenes as well as dual-focused raindrop degradations. Compared with conventional raindrop removal settings, this challenge places greater emphasis on robustness under diverse illumination conditions and focus patterns, making it more representative of practical applications. In total, 168 participants registered for the competition, and 17 teams finally submitted valid solutions and fact sheets for the testing phase. This report summarizes the challenge setting, evaluation protocol, participating methods, and final results. We hope this benchmark and the collected solutions can further promote research on real-world raindrop removal.

\section{Challenge}
\label{sec:challenge}
Following the first challenge~\cite{li2025ntire}, the NTIRE 2026 Second Challenge on Day and Night Raindrop Removal for Dual-Focused Images is still built upon the Raindrop Clarity dataset~\cite{jin2024raindrop}, which provides real-world samples covering both daytime and nighttime scenarios, as well as two representative focusing modes, namely raindrop-focused and background-focused images. 
The difference compared with the first challenge is that we adjusted the competition split by utilizing 14,139 images for training, 407 images for validation, and 593 images for testing. The training set follows the original benchmark release and serves as the basis for model development. The validation and testing sets, on the other hand, are organized specifically to keep the balance between different scenarios.

The objective of this challenge is not only to rank restoration methods on a common benchmark, but also to encourage the development of more robust and practically useful solutions for raindrop removal in the wild. By combining day/night scenes and dual-focused degradations into a unified benchmark, the challenge aims to provide a more realistic scheme for future research on raindrop removal under complex imaging conditions.

\subsection{Evaluation Protocol}

The restored results are assessed using three complementary image quality metrics, namely PSNR, SSIM, and LPIPS, which jointly reflect reconstruction fidelity and perceptual similarity. Following the evaluation setting adopted in the previous edition, the final ranking score is defined as
\begin{equation}
    \mathrm{Score} = 10\times\mathrm{PSNR (Y)} + 10\times \mathrm{SSIM (Y)} - 5\times \mathrm{LPIPS},
    \label{eq:finalscore}
\end{equation}
where higher PSNR and SSIM values lead to a better score, while a lower LPIPS value is preferred.

For PSNR and SSIM, the evaluation is conducted on the $\mathrm{Y}$ channel after converting the restored and reference images from the RGB color space to $\mathrm{YCbCr}$. For LPIPS, the image pixel values are first normalized to the range of $[-1,1]$, and then the perceptual distance is computed using the AlexNet-based configuration between the restored result and the corresponding ground-truth image.

\subsection{Competition Phases}
The competition is organized in two sequential stages on codabench~\cite{xu2022codabench}: a development stage for method validation and a testing stage for final ranking. In the first stage, participants are given access to the training data and a validation set without released ground truth, allowing them to train their models and obtain online feedback under the official metrics. For each submission, the platform returns the overall score together with PSNR, SSIM, and LPIPS, which helps participants assess both reconstruction fidelity and perceptual quality. For the development stage, we provide 4,713 triplets,  totaling 14,139 images for training and 407 rainy images for validation. Based on these data, participants can repeatedly refine their methods and submit restored results to the platform. A total of 637 submissions from 35 teams were received in this stage. The second stage serves as the final evaluation phase of the challenge. Participants are required to process 593 rainy images and submit their restored outputs for official scoring. The final ranking is determined by the score in Eq.~\ref{eq:finalscore}. Overall, 54 teams participated in the final testing stage. Among them, 17 teams submitted valid fact sheets and source codes, which were taken into account for the final ranking.

This challenge is one of the challenges associated with the NTIRE 2026 Workshop~\footnote{\url{https://www.cvlai.net/ntire/2026/}} on:
deepfake detection~\cite{ntire26deepfake}, 
high-resolution depth~\cite{ntire26hrdepth},
multi-exposure image fusion~\cite{ntire26raim_fusion}, 
AI flash portrait~\cite{ntire26raim_portrait}, 
professional image quality assessment~\cite{ntire26raim_piqa},
light field super-resolution~\cite{ntire26lightsr},
3D content super-resolution~\cite{ntire263dsr},
bitstream-corrupted video restoration~\cite{ntire26videores},
X-AIGC quality assessment~\cite{ntire26XAIGCqa},
shadow removal~\cite{ntire26shadow},
ambient lighting normalization~\cite{ntire26lightnorm},
controllable Bokeh rendering~\cite{ntire26bokeh},
rip current detection and segmentation~\cite{ntire26ripdetseg},
low light image enhancement~\cite{ntire26llie},
high FPS video frame interpolation~\cite{ntire26highfps},
Night-time dehazing~\cite{ntire26nthaze,ntire26nthaze_rep},
learned ISP with unpaired data~\cite{ntire26isp},
short-form UGC video restoration~\cite{ntire26ugcvideo},
raindrop removal for dual-focused images,
image super-resolution (x4)~\cite{ntire26srx4},
photography retouching transfer~\cite{ntire26retouching},
mobile real-word super-resolution~\cite{ntire26rwsr},
remote sensing infrared super-resolution~\cite{ntire26rsirsr},
AI-Generated image detection~\cite{ntire26aigendet},
cross-domain few-shot object detection~\cite{ntire26cdfsod},
financial receipt restoration and reasoning~\cite{ntire26finrec},
real-world face restoration~\cite{ntire26faceres},
reflection removal~\cite{ntire26reflection},
anomaly detection of face enhancement~\cite{ntire26anomalydet},
video saliency prediction~\cite{ntire26videosal},
efficient super-resolution~\cite{ntire26effsr},
3d restoration and reconstruction in adverse conditions~\cite{ntire26realx3d},
image denoising~\cite{ntire26denoising},
blind computational aberration correction~\cite{ntire26aberration},
event-based image deblurring~\cite{ntire26eventblurr},
efficient burst HDR and restoration~\cite{ntire26bursthdr},
low-light enhancement: `twilight cowboy'~\cite{ntire26twilight},
and efficient low light image enhancement~\cite{ntire26effllie}.

\begin{table*}[tp]
    \centering
    \caption{Quantitative results of the NTIRE~2026 The Second Challenge on Day and Night Raindrop Removal for Dual?Focused Images.  The best and second best values per column are highlighted in \textcolor{red}{red} and \textcolor{blue}{blue}, respectively.  "Params." and "GFlops" are reported from the teams’ factsheets.  A checkmark ($\checkmark$) indicates the use of ensembles or extra data, while a cross ($\otimes$) indicates otherwise.}
    \resizebox{\textwidth}{!}{
    \begin{tabular}{cc|cccc|cc|c|c|c}
    \toprule
    Team & Leader & Final Score $\uparrow$ & PSNR$\uparrow$ & SSIM$\uparrow$ & LPIPS $\downarrow$ & Params.\,(M) & GFlops\,(G) & Ensemble & Extra Data & Rank\\ \midrule
    AIIA\textendash Lab & Runzhe~Li & \textcolor{red}{35.2378} & \textcolor{red}{28.3392} & \textcolor{red}{0.8265} & 0.2732 & 16.6 & 129.9 & $\checkmark$ & $\otimes$ & 1\\
    raingod & Hongde~Gu & \textcolor{blue}{35.2186} & \textcolor{blue}{28.2817} & \textcolor{blue}{0.8255} & \textcolor{blue}{0.2636} & 16.6 & 129.9 & $\checkmark$ & $\checkmark$ & 2\\
    BUU\_CV & Qiyu~Rong & 35.0360 & 28.1469 & 0.8222 & 0.2665 & 26.89 & \textcolor{blue}{42.33} & $\checkmark$ & $\checkmark$ & 3\\
    RetinexDualV2 & Mohab~Kishawy & 33.8556 & 27.2379 & 0.8061 & 0.2887 & 4.8 & 301.6 & $\checkmark$ & $\otimes$ & 4\\
    ULR & Anh\textendash Kiet~Duong & 33.7505 & 27.0572 & 0.7966 & \textcolor{red}{0.2547} & 593 & 8396 & $\checkmark$ & $\otimes$ & 5\\
    GU\textendash day~Mate & Wangzhi~Xing & 32.9288 & 26.5547 & 0.7826 & 0.2903 & \textcolor{red}{2.14} & 64.2 & $\checkmark$ & $\otimes$ & 6\\
    Derain & Wenjie~Li & 32.7158 & 26.5915 & 0.7882 & 0.3515 & 16.1 & 132.3 & $\otimes$ & $\otimes$ & 7\\
    NTR & Jiachen~Tu & 32.1273 & 26.1084 & 0.7674 & 0.3310 & 142.5 & - & $\checkmark$ & $\otimes$ & 8\\
    Cidaut~AI & Paula~Garrido~Mellado & 31.9468 & 25.8394 & 0.7653 & 0.3092 & \textcolor{blue}{2.95} & \textcolor{red}{13.26} & $\otimes$ & $\otimes$ & 9\\
    NCHU\textendash CVLab & Zhidong~Zhu & 31.9438 & 25.8288 & 0.7640 & 0.3050 & 29.38 & 318 & $\otimes$ & $\otimes$ & 10\\
    NUDT\textendash Deeplter & Zida~Zhang & 31.9235 & 25.8568 & 0.7663 & 0.3193 & 26.1 & 140 & $\checkmark$ & $\otimes$ & 11\\
    DGLTeam & Guanglu~Dong & 31.8563 & 25.7230 & 0.7682 & 0.3098 & 42.6 & - & $\checkmark$ & $\otimes$ & 12\\
    MMAIrider & Li\textendash Wei~Kang & 31.6923 & 25.7243 & 0.7655 & 0.3373 & 10.17 & - & $\checkmark$ & $\otimes$ & 13\\
    Rain\textendash SVNIT & Abhishek~Rajak & 31.6514 & 25.8245 & 0.7638 & 0.3623 & 26.1 & 140 & $\checkmark$ & $\otimes$ & 14\\
    Just~JiT & Huilin~Zhao & 31.2484 & 25.3451 & 0.7601 & 0.3396 & 953 & 182 & $\checkmark$ & $\checkmark$ & 15\\
    PSU & Bilel~Benjdira & 31.1371 & 25.4649 & 0.7519 & 0.3693 & 45.6 & 493.36 & $\otimes$ & $\otimes$ & 16\\
    BITssvgg & Hao~Yang & 30.9433 & 25.1389 & 0.7495 & 0.3382 & 16.87 & 496 & $\otimes$ & $\checkmark$ & 17\\
    \bottomrule
    \end{tabular}}
    \label{tab:results}
\end{table*}

\section{Challenge Results}
We have summarized the challenge results in Table~\ref{tab:results}. 
Team AIIA--Lab achieved the best overall performance in the challenge, ranking 1st with a final score of 35.2378. It also obtained the best PSNR (28.3392) and SSIM (0.8265), together with a competitive LPIPS of 0.2732, while using 16.6M parameters and 129.9 GFlops without extra data. The 2nd-place team, raingod, followed closely with a final score of 35.2186, benefiting from strong PSNR (28.2817), SSIM (0.8255), and the second-best LPIPS (0.2636). The 3rd-place team, BUU\_CV, obtained a final score of 35.0360, with 28.1469 PSNR, 0.8222 SSIM, and 0.2665 LPIPS, while requiring only 42.33 GFlops, indicating a strong balance between performance and efficiency.

In addition to the leading teams, the benchmark highlights diverse trade-offs among restoration quality, perceptual quality, and model complexity. For instance, Team ULR achieved the best LPIPS (0.2547), suggesting superior perceptual restoration capability, whereas lightweight methods such as teams GU--day Mate and Cidaut AI demonstrated competitive efficiency with only 2.14M and 2.95M parameters, respectively. Overall, the submitted solutions exhibit a wide range of model designs, from high-performing deraining systems to compact architectures, reflecting the richness and practical relevance of this challenge.
\section{Teams and Methods}
\label{sec:teams_and_methods}

% The following subsections are ordered according to the final ranking of the challenge.

\subsection{AIIA-Lab}

\begin{figure}[t]
    \centering
    \includegraphics[width=1.0\linewidth]{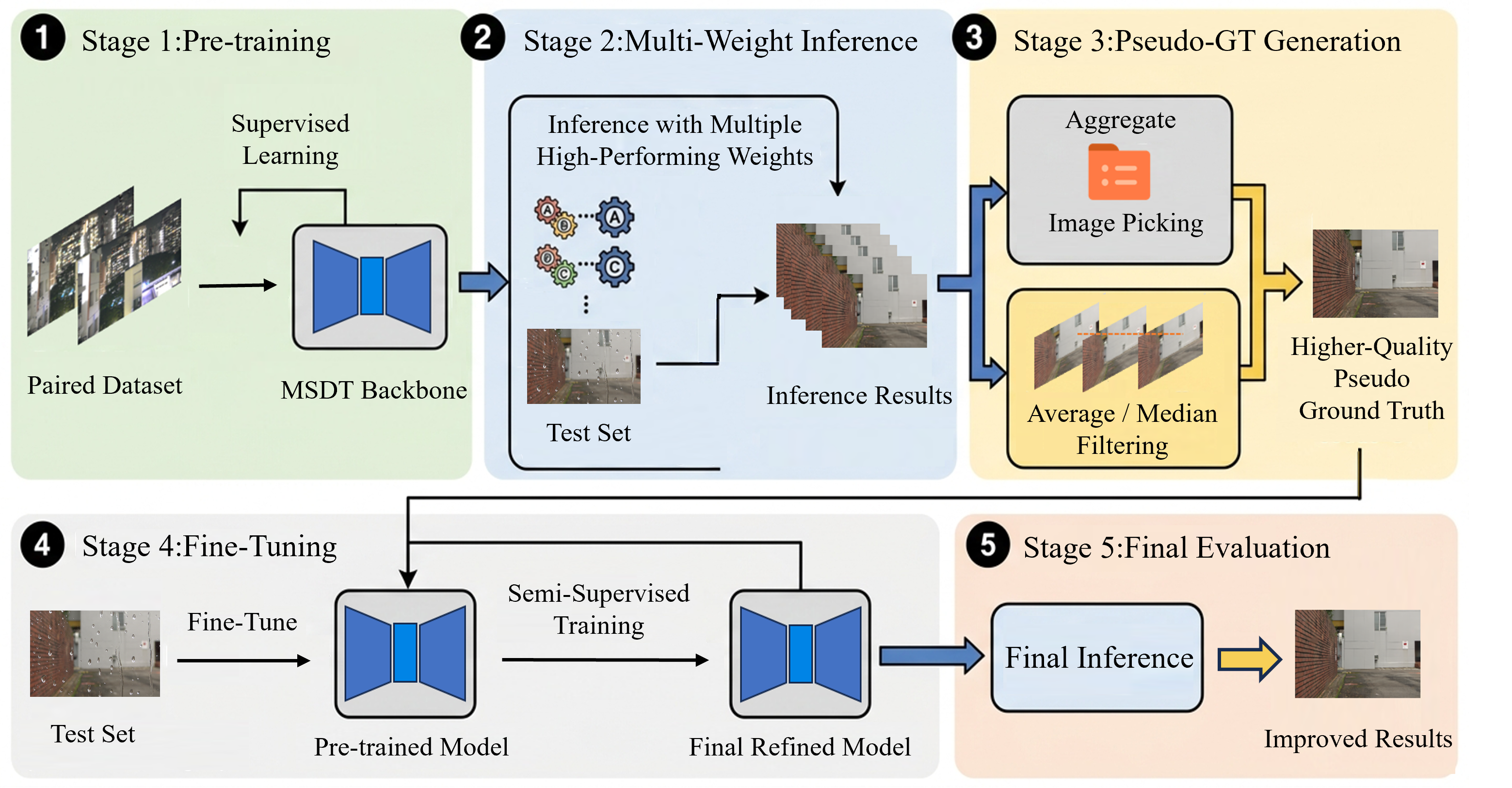}
    \caption{The pipeline of the method proposed by Team AIIA-Lab}
    \label{fig:AIIA_Lab}
\end{figure}

This team proposes a challenge-oriented multi-stage pipeline for dual-focused day/night raindrop removal, as illustrated in Fig.~\ref{fig:AIIA_Lab}. 
They first compared several representative restoration backbones, including Restormer~\cite{zamir2022restormer_deblur-transformer}, Histoformer~\cite{sun2024restoringhistogram}, and MSDT~\cite{chen2024rethinkingMSDT}, and selected MSDT as the final model based on its superior validation performance and optimization potential. 

Building upon this backbone, they optimized the model through hyperparameter adjustment and long-term training, and retained several strong checkpoints. These checkpoints were then used for checkpoint-based result selection across different scenes. To further exploit scene-level consistency, outputs from the same scene were fused using mean or median aggregation to construct pseudo-GTs, followed by an additional refinement stage. The final performance is achieved through the joint effect of backbone selection, checkpoint-based result selection, scene-level fusion, and pseudo-GT-based refinement.

\noindent\textbf{Training Details.} 
The training consists of two stages. In the first stage, the MSDT backbone~\cite{chen2024rethinkingMSDT} is trained with optimized hyperparameters under mixed degradations of day/night and dual-focused settings, using standard augmentations such as random cropping, flipping, and rotation. Multiple strong checkpoints are retained during long-term training. In the second stage, pseudo ground truth is constructed via scene-level fusion and used to fine-tune the model with a smaller learning rate and a more conservative optimization strategy.

\noindent\textbf{Testing Details.} 
During testing, multiple checkpoints are used to generate candidate restorations, and the most suitable results are selected for each scene. Scene-level mean or median fusion is further applied to enhance consistency and reduce artifacts. The final results are obtained from the best-performing checkpoint after refinement.

\noindent\textbf{Implementation Details.} 
The method is implemented in PyTorch with MSDT~\cite{chen2024rethinkingMSDT} as the backbone. The model is trained using Adam with a cosine annealing learning rate schedule, for 200 epochs on $256\times256$ patches with a batch size of 1 on a single GPU. No external data is used.

\subsection{raingod}

\begin{figure}[t]
    \centering
    \begin{subfigure}{0.49\linewidth}
        \includegraphics[width=\linewidth]{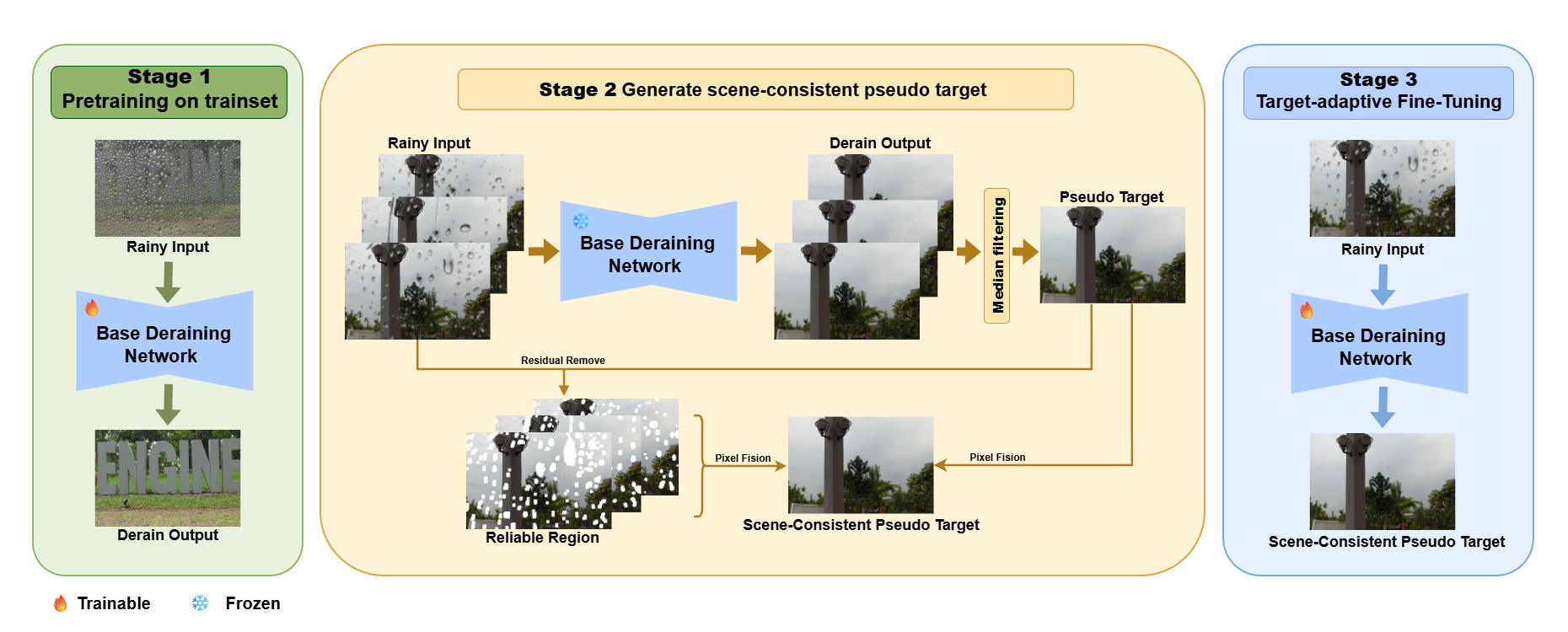}
        \caption{Pipeline}
    \end{subfigure}
    \hfill
    \begin{subfigure}{0.49\linewidth}
        \includegraphics[width=\linewidth]{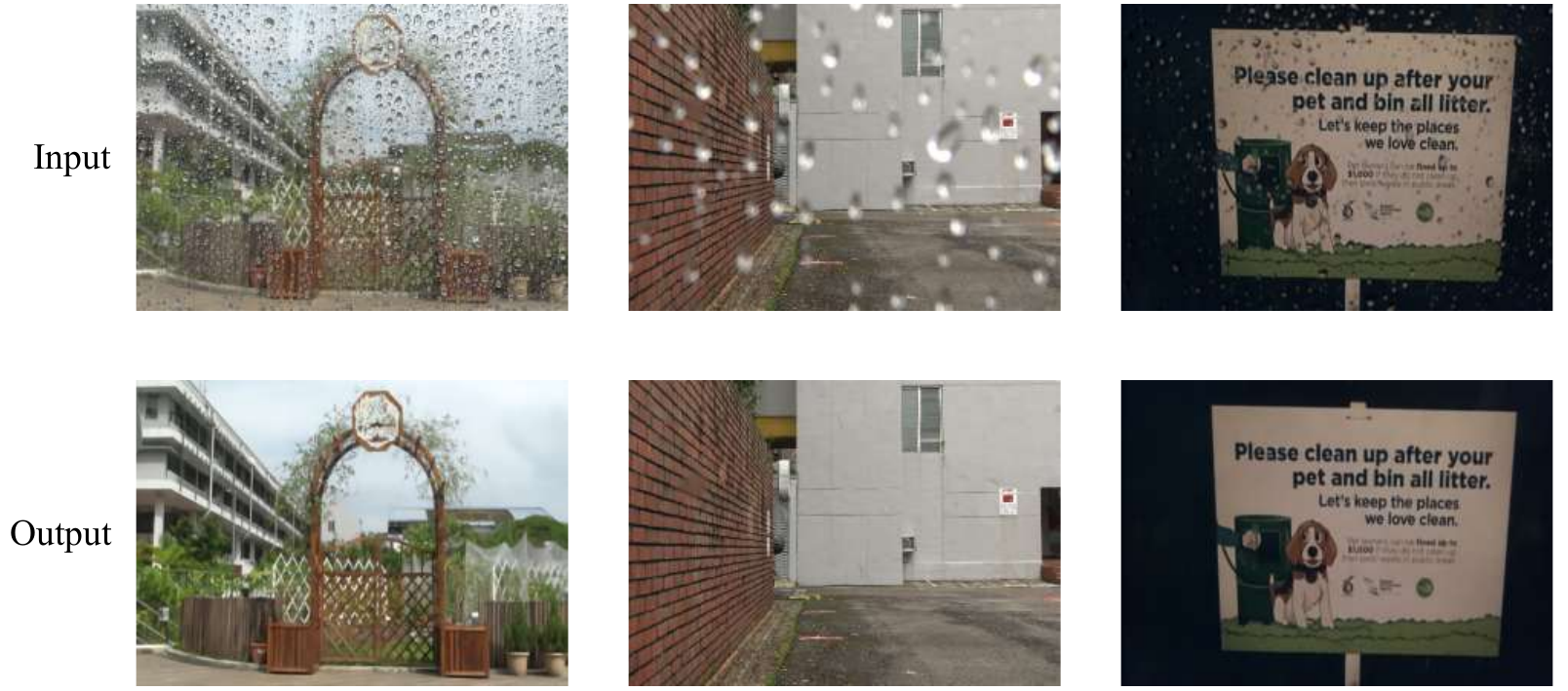}
        \caption{Restoration results}
    \end{subfigure}
    \caption{Method overview and qualitative results of Team raingod}
    \label{fig:raingod}
\end{figure}

This team proposes a three-stage framework for dual-focused day and night raindrop removal, as shown in Fig.~\ref{fig:raingod}. 
In the first stage, considering the effectiveness of multi-scale modeling in image deraining, they pretrain the MSDT model~\cite{chen2024rethinkingMSDT} on the RainDrop Clarity dataset~\cite{jin2024raindrop}, and further introduce the UAV-Rain1k dataset~\cite{chang2024uavrain1k} to enhance generalization. 

In the second stage, preliminary restorations are generated for each scene, and scene‑consistent pseudo targets are constructed. Specifically, median filtering is applied to obtain coarse targets, which are then combined with reliable clean regions extracted from the rainy inputs to form refined pseudo supervision. In the third stage, the model is fine‑tuned using the generated pseudo targets to better adapt to the target scenes. During testing, the refined model is used for inference, followed by a median‑filter‑based refinement module to further improve results.

\subsection{BUU\_CV}

\begin{figure}[t]
    \centering
    \includegraphics[width=1.0\linewidth]{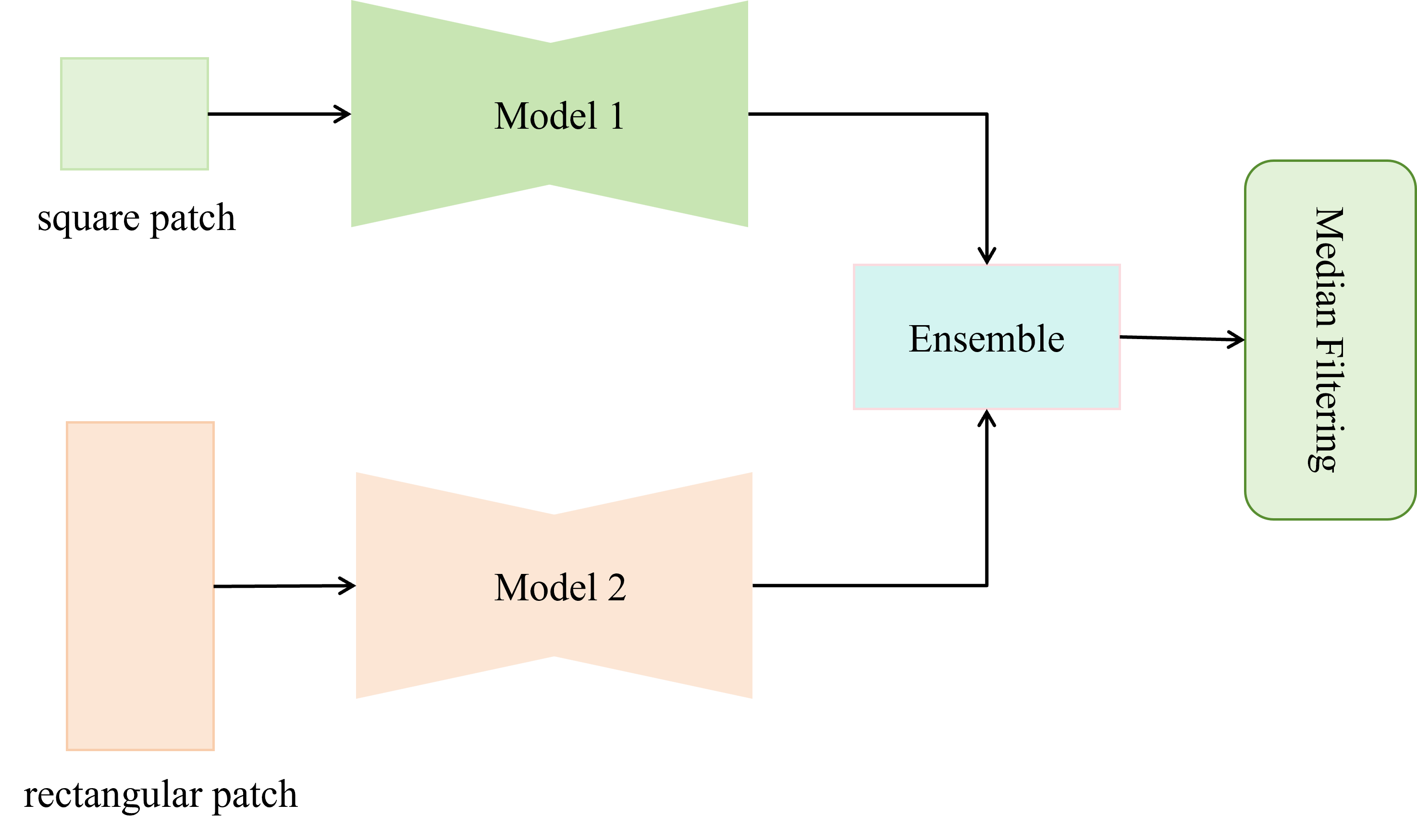}
    \caption{The pipeline of the method proposed by Team BUU\_CV}
    \label{fig:BUU_CV}
\end{figure}

This team improves the modeling of elongated raindrops by combining two complementary models, as illustrated in Fig.~\ref{fig:BUU_CV}. 
Specifically, they employ STRRNet~\cite{rong2025STRRNet} and Restormer~\cite{zamir2022restormer_deblur-transformer}. STRRNet is trained with rectangular patches to better model directional streak-like raindrops, while Restormer is trained with square patches to better preserve local textures and fine details. Their outputs are fused, followed by a median filtering step to further suppress residual artifacts and improve visual quality.

\noindent\textbf{Training Details.} 
The models are trained using the AdamW optimizer with an initial learning rate of $3\times10^{-4}$. The training loss consists of $\mathcal{L}_1$ loss and SSIM loss. Geometric data augmentations are adopted to improve generalization. The batch size is set to 4, and the total number of iterations is 200,000 on a single RTX 4090 GPU.

\noindent\textbf{Testing Details.} 
During inference, sliding‑window processing is applied with two models trained under different patch settings. Their outputs are combined using a weighted ensemble strategy to produce the final result.

\noindent\textbf{Implementation Details.} 
The method is implemented using two independently trained models with different patch configurations. The rectangular-patch model focuses on elongated raindrop removal, while the square-patch model enhances texture preservation. No extra data is used.

\subsection{RetinexDualV2}

\begin{figure}[t]
    \centering
    \includegraphics[width=1.0\linewidth]{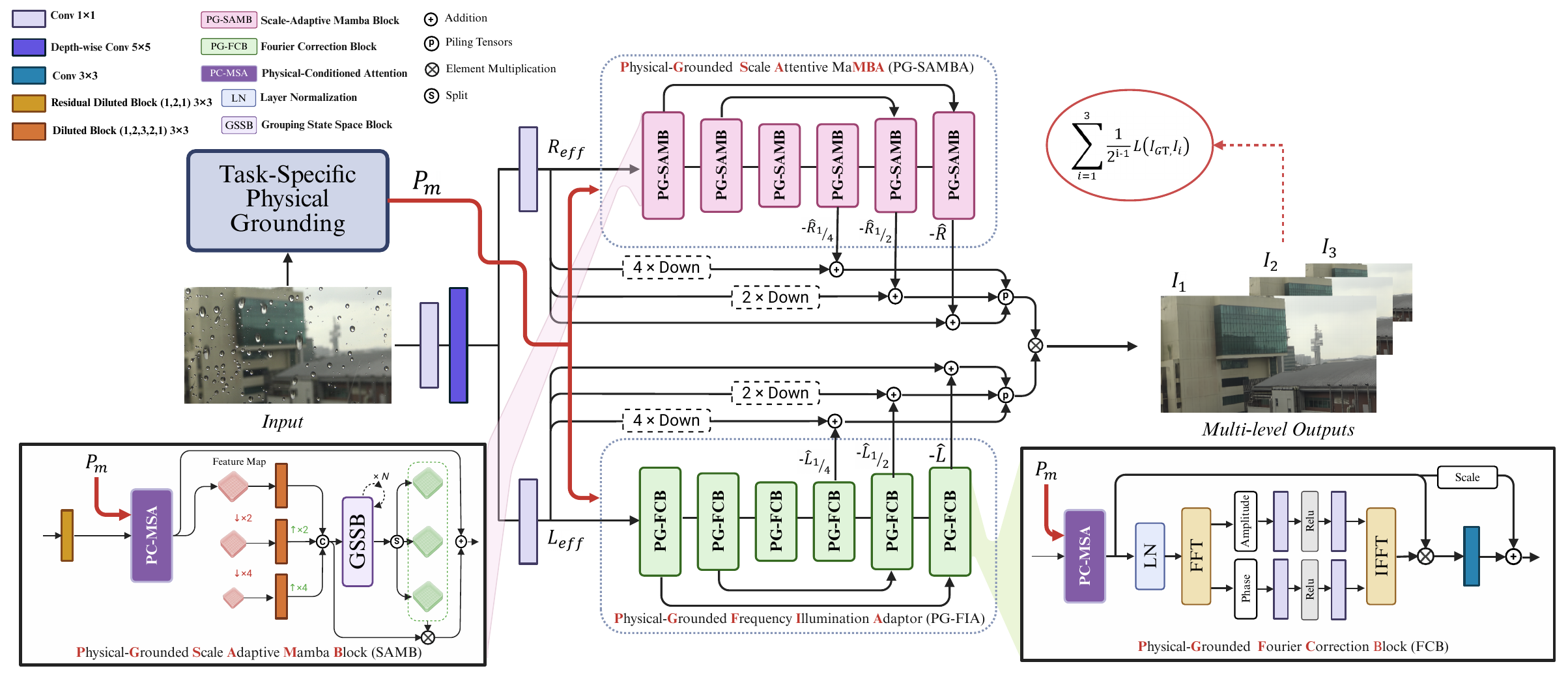}
    \caption{The pipeline of the method proposed by Team RetinexDualV2}
    \label{fig:RetinexDualV2}
\end{figure}

RetinexDualV2 \cite{kishawy2026retinexdualv2physicallygroundeddualretinex}, following its predecessor RetinexDual \cite{kishawy2025retinexdualretinexbaseddualnature}, proposes a Retinex-based method that recognizes the dual nature of the Retinex decomposition by implementing two distinct complementary branches, each built upon the characteristics of the reflectance and illumination components. The main contribution is the integration of a physical grounding prior, which is derived from the residual rain intensity mask computed as the difference between the rainy image and a blurred version. A shallow UNet $\Phi_{\text{rain}}$ was trained on paired drops and the residual rain intensity mask from the training set to estimate the rain intensity in the testing set. This mask is injected into the network via a Physical-Grounding multi-head attention mechanism in a Scale Attentive Mamba branch and a Frequency Illumination adaptor branch. After initial supervised training, the model is fine-tuned on pseudo ground truths obtained by soft blending across all images from the same scene using the residual rain mask as weights. Finally, outputs from different scene images are fused using soft blending to produce the final restoration.

\noindent\textbf{Training Details.} 
RetinexDualV2 is trained in a two-stage manner on $768\times768$ random patches from the provided training data. The first stage uses direct paired supervision with a multi-scale loss combining Charbonnier, SSIM, FFT and perceptual terms, optimized with the AdamW optimizer (initial learning rate $1\times10^{-5}$, weight decay $0.001$) and CosineAnnealingRestart scheduling. Geometric augmentations and MixUp (beta=1.2) are applied. In the second stage, the model is fine-tuned on pseudo ground truths generated via weighted averaging based on the residual rain intensity mask to better handle rain variability. Training spans several hundred epochs on a single GPU, and no external data are used.

\noindent\textbf{Testing Details.} 
Inference operates on full-resolution inputs padded to multiples of 128, processing the image and its rain mask in a single forward pass. The output is cropped back to the original dimensions. A soft blend across all images in the same scene, weighted by the residual rain mask, is then applied to produce the final result.

\noindent\textbf{Implementation Details.} 
The architecture contains about $4.8\,$M parameters and has a computational complexity of roughly $301.6\,$GFLOPs per $768\times768$ input. Training and inference are implemented in PyTorch using the BasicSR framework. No external datasets are used, and the only ensemble strategy is the soft blending across scene images.

\subsection{ULR}

\begin{figure}[t]
    \centering
    \includegraphics[width=1.0\linewidth]{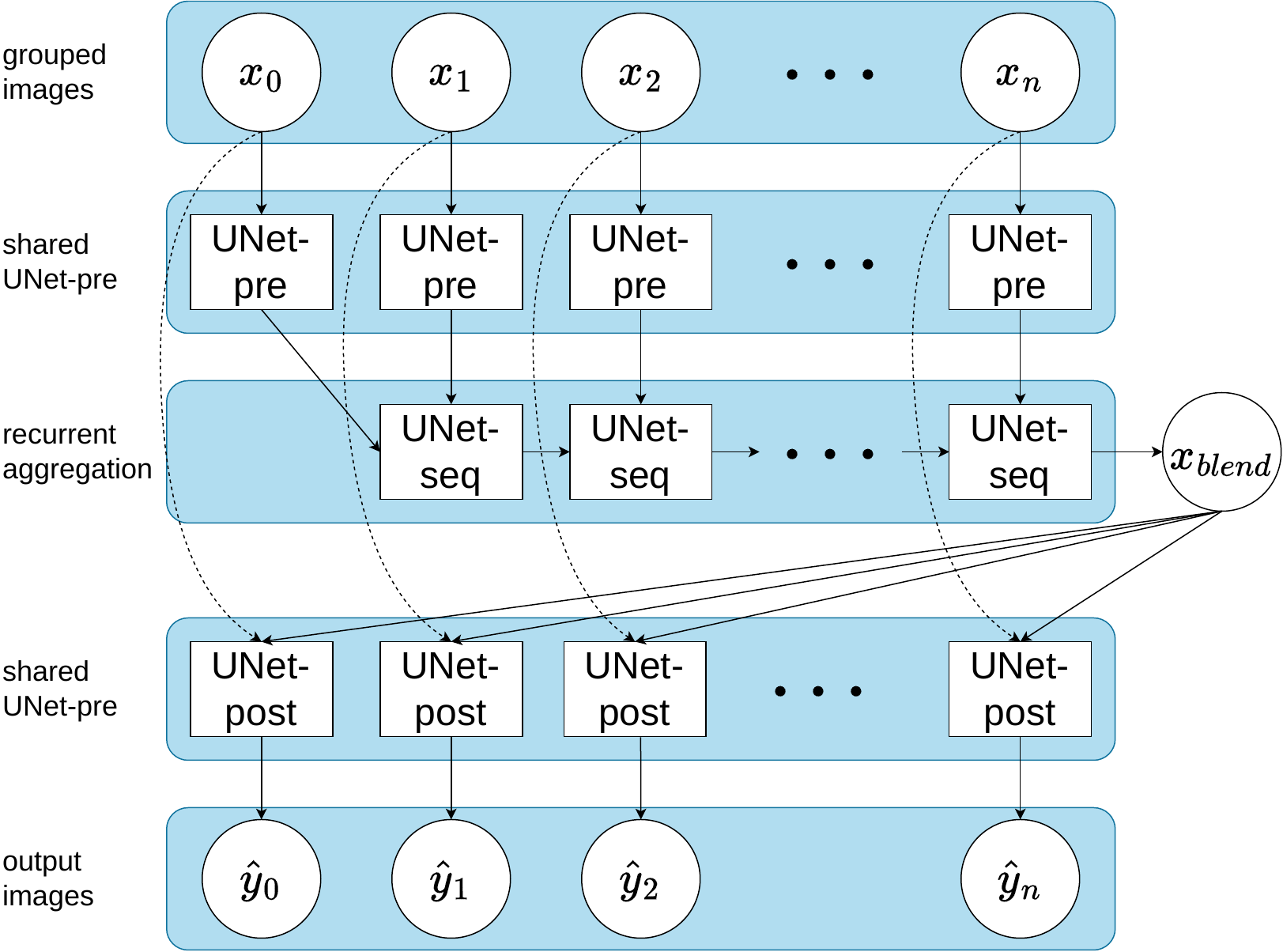}
    \caption{The pipeline of the method proposed by Team ULR}
    \label{fig:ULR}
\end{figure}

This team proposes a three-stage sequential framework for raindrop removal, as illustrated in Fig.~\ref{fig:ULR}. The method consists of a preprocessing stage, a recurrent aggregation stage, and a postprocessing stage. In the first stage, a UNet (UNet-pre) processes each input image independently to produce preliminary restorations~\cite{unet}. In the second stage, a recurrent UNet (UNet-seq) takes images from the same scene as input and generates an aggregated representation. In the final stage, another UNet (UNet-post) takes both the aggregated image and the original input image to produce the final restored result.

\noindent\textbf{Training Details.} 
The three components—UNet-pre, UNet-seq and UNet-post—are trained sequentially for 100 epochs on the provided dataset. Training uses the Adam optimizer with a learning rate of $1\times10^{-6}$, a batch size of 4 and no external data.

\noindent\textbf{Testing Details.} 
During inference, images are grouped by scene, and the model processes them at the original resolution. A mean ensemble of outputs from different models is applied to obtain the final results.

\noindent\textbf{Implementation Details.} 
The model is trained for 100 epochs with a batch size of 4 using the Adam optimizer and a learning rate of $1\times10^{-6}$ on a single NVIDIA H100 GPU. No external data are used.

\subsection{GU-day Mate}

\begin{figure}[t]
    \centering
    \includegraphics[width=1.0\linewidth]{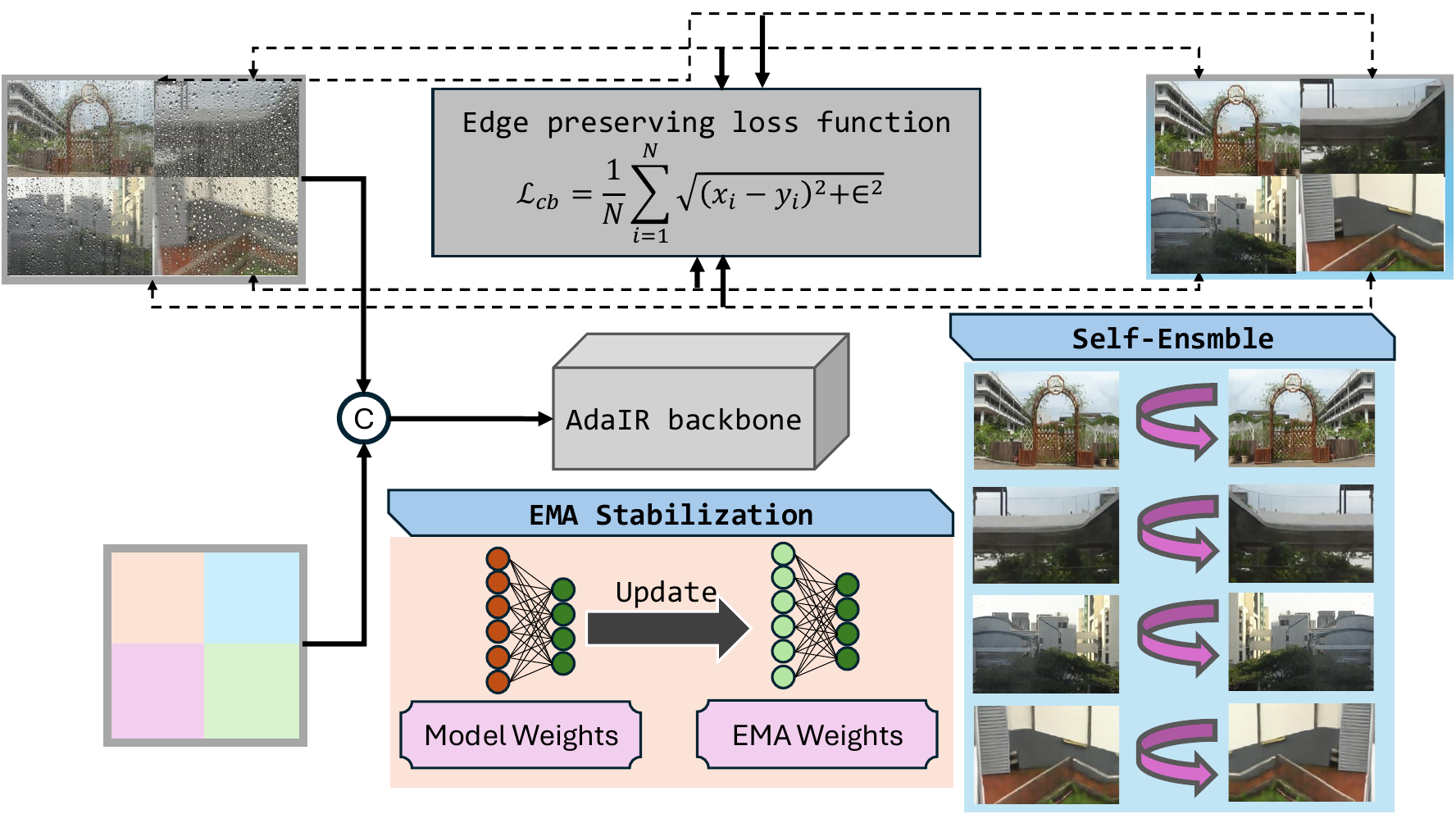}
    \caption{The pipeline of the method proposed by Team GU-day Mate}
    \label{fig:GU_day_Mate}
\end{figure}

This team proposes a frequency-aware transductive fine-tuning framework built upon the AdaIR architecture~\cite{cui2025adair}, as shown in Fig.~\ref{fig:GU_day_Mate}. The method follows a multi-stage pipeline designed to bridge the distribution gap between training and test data.

Specifically, the approach introduces a category-aware pseudo ground-truth (Pseudo-GT) construction strategy. For blurry samples, handcrafted unsharp masking is used to preserve edge structures, while for other degradations, Pseudo-GTs are generated from test-time augmented AdaIR outputs. These pseudo labels are then used for transductive fine-tuning directly on the test scenes.

To further enhance robustness, the method adopts a dynamic training curriculum with progressively increasing resolutions (from $128$ to $512$), combined with frequency-domain MixUp  to decouple semantic phase information from degradation amplitude. Model stability is improved using exponential moving average (EMA) during optimization.

\noindent\textbf{Training Details.} 
Training follows a dynamic curriculum over approximately 120 epochs, progressively increasing crop sizes from $128\times128$ to $512\times512$ while reducing the batch size. The AdaIR backbone is optimized with the AdamW optimizer (weight decay $0.01$) and an initial learning rate of $2\times10^{-4}$, which decays to $1\times10^{-6}$ via a cosine annealing schedule. A hybrid objective combining Charbonnier and Sobel-gradient losses is used, and frequency-domain MixUp and EMA further stabilize training. No external data are employed.

\noindent\textbf{Testing Details.} 
For each test scene, the model performs a brief transductive fine-tuning using the pseudo-GTs constructed from the scene itself, then applies an eight-fold geometric self-ensemble~\cite{timofte2016seven}. The ensemble results are averaged to obtain the final restoration.

\noindent\textbf{Implementation Details.} 
The AdaIR-based solution contains about $2.14\,$M parameters and requires roughly $64.2\,$GFLOPs per $512\times512$ crop. Training takes around 24 hours on an 8$\times$A100 GPU cluster. No extra datasets are used.

\subsection{Derain}

This team proposes a semantic-constrained transformer framework based on a Restormer architecture~\cite{zamir2022restormer_deblur-transformer} with a dual-guidance strategy. The method introduces two disentangled semantic priors generated by a multimodal large language model (MLLM), namely a degradation embedding (DE) and a clean embedding (CE), to separately model degradation removal and content reconstruction.

Specifically, the encoder incorporates Degradation Fusion of Experts (DFE) modules, which are adaptively modulated by the degradation embedding to enhance feature extraction from degraded inputs. The clean embedding is then injected into the bottleneck through a cross-attention mechanism, guiding semantically consistent feature refinement. The decoder finally reconstructs the restored image based on the refined representation.

\noindent\textbf{Training Details.} 
The semantic‑constrained transformer is trained from scratch using the Adam optimizer with $\beta_1=0.9$ and $\beta_2=0.99$.  Training runs for roughly $500\,000$ iterations on $192\times192$ patches with a batch size of~8.  The learning rate starts at $5\times10^{-4}$ and decays to $1\times10^{-5}$ via a cosine schedule.  A mixed loss comprising $\ell_1$ and Fourier terms (weights~1 and~0.1) guides optimization, and no extra datasets are used.

\noindent\textbf{Implementation Details.} 
The network, built on a Restormer backbone with dual semantic priors, has approximately $16.1$~million parameters and a computational cost of about $132.3\,$GFLOPs.  It is trained on a single RTX~4090 GPU for around one day and does not rely on self‑ensembles or external data.

\subsection{NTR}

\begin{figure}[t]
    \centering
    \includegraphics[width=1.0\linewidth]{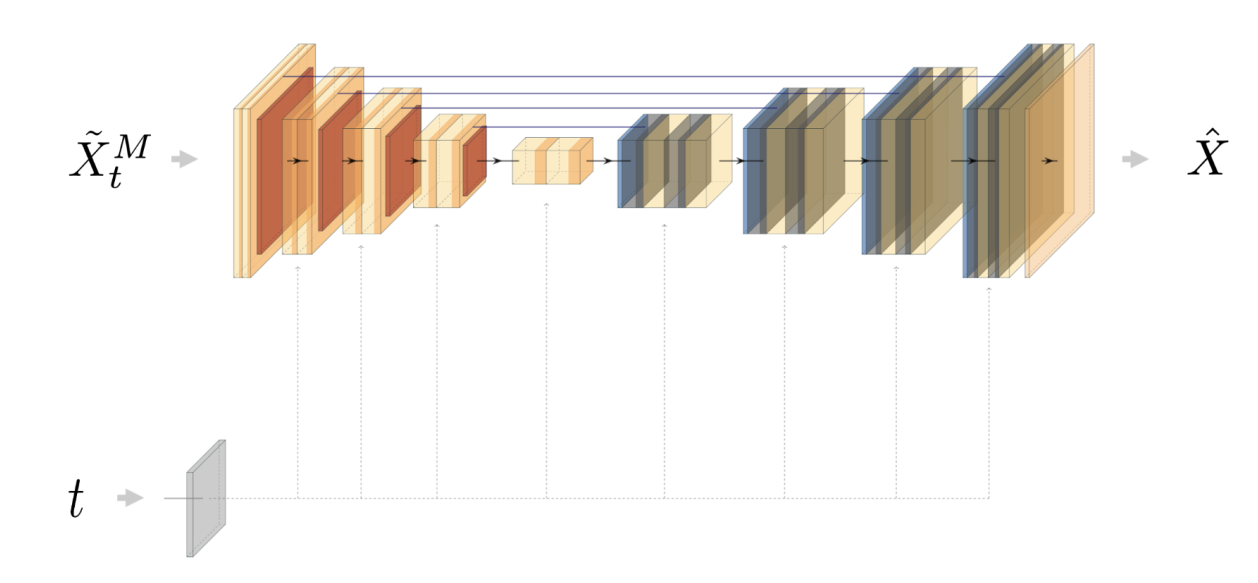}
    \caption{The pipeline of the method proposed by Team NTR}
    \label{fig:NTR}
\end{figure}

This team proposes a diffusion-based restoration frameworkcwith a two-stage training paradigm, as illustrated in Fig.~\ref{fig:NTR}, combining masked diffusion autoencoding pretraining and supervised score-based fine-tuning.

In the first stage, the model is pretrained using Masked Diffusion Autoencoding (MDAE)~\cite{he2022mae,tu2025scoremri}, where each input is corrupted by both spatial masking and noise injection. 

In the second stage, the pretrained model is fine-tuned using a score-based objective (D2S-SFT)~\cite{tu2025d2ssft}, where the model predicts clean images from noisy degraded inputs at a fixed noise level.

The network architecture is based on a time-conditioned diffusion transformer~\cite{tu2025scoremri}, implemented as a U-Net encoder-decoder operating at multiple scales.

\subsection{Cidaut AI}

\begin{figure}[t]
    \centering
    \includegraphics[width=1.0\linewidth]{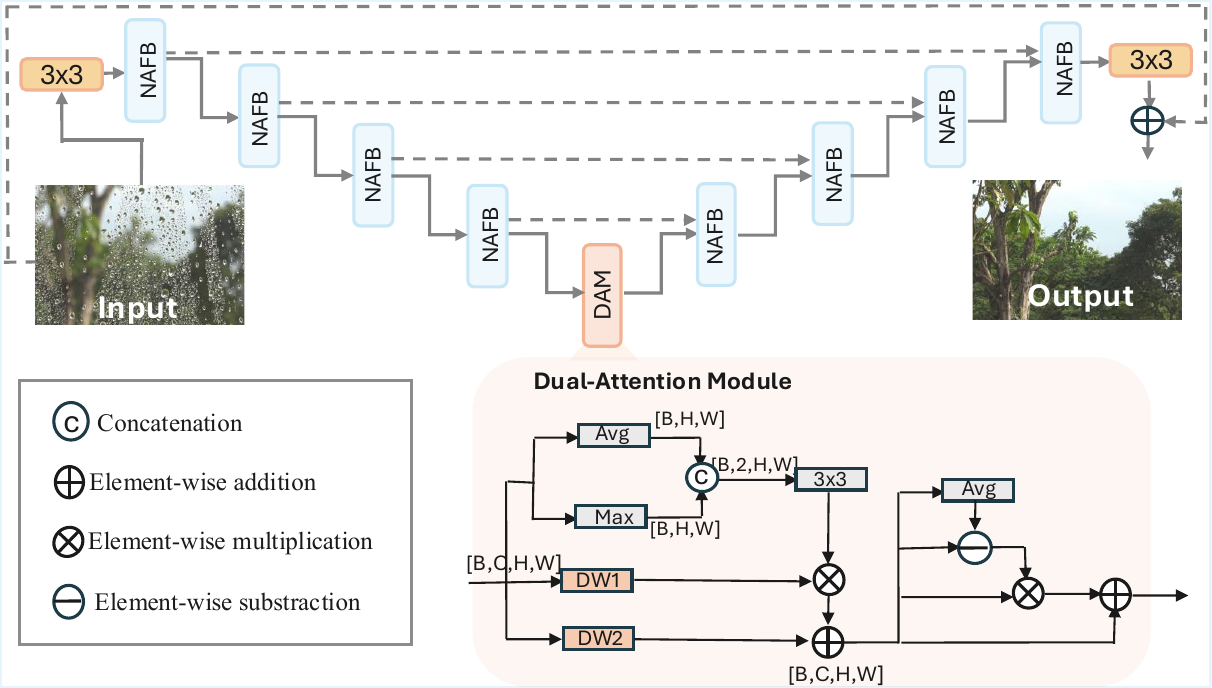}
    \caption{The pipeline of the method proposed by Team Cidaut AI}
    \label{fig:CidautAI}
\end{figure}

This team proposes a raindrop removal framework named DualDrop, built upon a NAFNet~\cite{chen2022simpleNAFNet} backbone with a U-Net~\cite{unet} architecture, as illustrated in Fig.~\ref{fig:CidautAI}. The network consists of four encoder and decoder stages, where each stage is composed of multiple NAFBlocks to balance restoration performance and computational efficiency.

The core component is a Dual-Attention Module (DAM) introduced at the bottleneck. This module combines spatial and frequency attention to enhance feature discrimination between raindrop-corrupted and clean regions. It adopts two parallel branches: one branch applies cascaded depthwise dilated convolutions with different receptive fields, while the other uses a standard depthwise convolution. The fused features are used to guide raindrop localization and restoration.

\noindent\textbf{Training Details.} 
DualDrop is trained for 400~epochs on the Raindrop Clarity dataset using $384\times384$ crops and a batch size of~16 on four NVIDIA H100 GPUs.  The AdamW optimizer with $\beta_1=\beta_2=0.9$ and a weight decay of $1\times10^{-3}$ is employed.  The learning rate begins at $2\times10^{-3}$ and decays to $1\times10^{-6}$ following a cosine annealing schedule.  No extra datasets or multi‑stage training are used.

\noindent\textbf{Implementation Details.} 
The single end‑to‑end model contains about $2.95\,$M parameters with a computational cost of roughly $13.26\,$GFLOPs.  It achieves a runtime of around $19\,$ms per $480\times720$ image on H100 GPUs and does not employ any ensemble or fusion strategies.

\subsection{NCHU-CVLab}

This team adopts a U-shaped encoder-decoder architecture based on the Histogram Transformer~\cite{sun2024restoringhistogram} for dual-focused raindrop removal. The network extracts shallow features through an overlapping convolutional embedding layer and processes them via multi-scale encoder and decoder stages connected with skip connections.

The core building block is the Histogram Transformer Block (HTB), which integrates dynamic-range histogram self-attention and a dual-scale gated feed-forward module. The histogram self-attention rearranges features according to intensity distributions and performs attention through bin-wise and frequency-wise representations, enabling interactions among spatially distant but similarly degraded pixels. The gated feed-forward module further enhances feature representation using multi-scale convolutional operations. Hierarchical downsampling and upsampling are employed to aggregate coarse-to-fine restoration features.

\noindent\textbf{Training Details.} 
Training adopts a progressive curriculum over five stages.  Patch sizes are increased from $128$ to $288$ pixels with corresponding iteration counts of 200k, 160k, 120k, 80k and 40k, while the mini‑batch per GPU is adjusted as 4, 2, 2, 1 and 1.  The AdamW optimizer (initial learning rate $3\times10^{-4}$, weight decay $1\times10^{-4}$) is used with a cosine‑annealing restart schedule decaying to $1\times10^{-6}$.  In total, about 600k iterations (approximately 754~epochs) are trained on four GPUs using random flipping and rotation augmentations.  No external data are involved.

\noindent\textbf{Testing Details.} 
The model follows a standard full‑resolution inference pipeline without ensemble strategies.

\noindent\textbf{Implementation Details.} 
The Histogram Transformer–based network has roughly $29.38\,$M parameters and about $318\,$GFLOPs complexity.  It processes a $480\times720$ image in around $1.2\,$s on four RTX~3090 GPUs.  Only the official training data are used.

\subsection{NUDT-Deeplter}

\begin{figure}[t]
    \centering
    \includegraphics[width=1.0\linewidth]{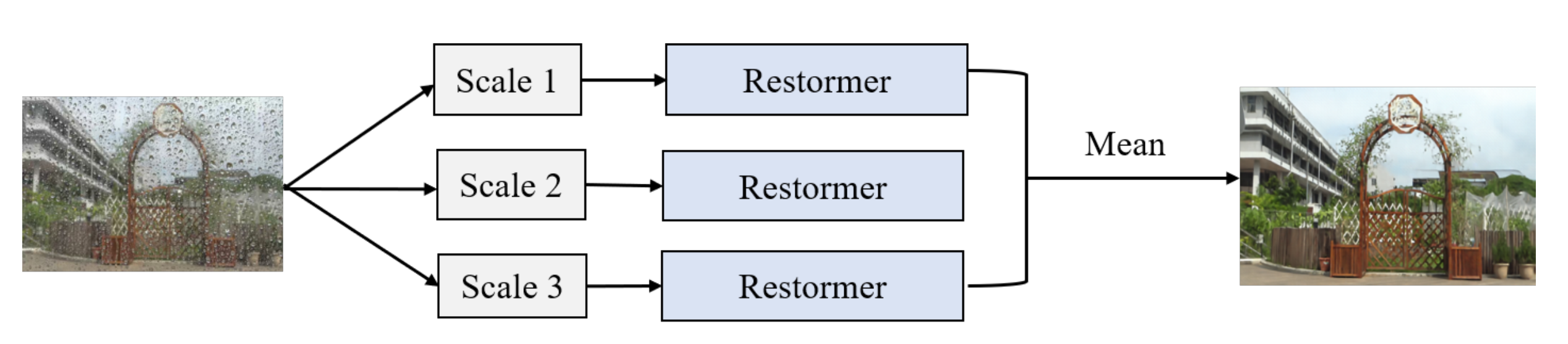}
    \caption{The pipeline of the method proposed by Team NUDT–Deeplter}
    \label{fig:NUDT_Deeplter}
\end{figure}

This team introduces a multi-scale test-time fusion framework built upon a Restormer backbone~\cite{zamir2022restormer_deblur-transformer}, as shown in Fig.~\ref{fig:NUDT_Deeplter}. At inference, images are resized to three different scales via bicubic interpolation to capture rain artefacts of varying sizes. Each scaled image is processed by a Restormer model using overlapping sliding windows, generating three intermediate restorations. These are then realigned to the original resolution and averaged to form the final output, effectively fusing complementary information from multiple resolutions.

\noindent\textbf{Training Details.} 
The Restormer backbone is trained using $512\times512$ crops randomly sampled from the competition data. Standard augmentations, including random flips and rotations, are applied. The loss combines $\ell_1$ reconstruction and a VGG-based perceptual term to balance fidelity and perceptual quality. Training is performed for 100~epochs on a single GPU with the Adam optimizer and a fixed learning rate. No external datasets are used.

\noindent\textbf{Testing Details.} 
During inference, the input image is resized to three scales (e.g., $1.0\times$, $0.8\times$, and $1.2\times$) and processed by the trained Restormer on sliding windows with overlap to handle large images. The outputs are aligned to the original scale using bicubic interpolation and averaged pixel-wise. This multi-scale fusion acts as a simple ensemble strategy that improves robustness without additional training.

\noindent\textbf{Implementation Details.} 
The approach uses a single Restormer model with approximately $26.1$~M parameters and a computational complexity of about $140\,$GFLOPs per $256\times256$ patch. No additional data are employed beyond the competition dataset. The three-scale fusion constitutes the only ensemble technique.

\subsection{DGLTeam}

\begin{figure}[t]
    \centering
    \includegraphics[width=1.0\linewidth]{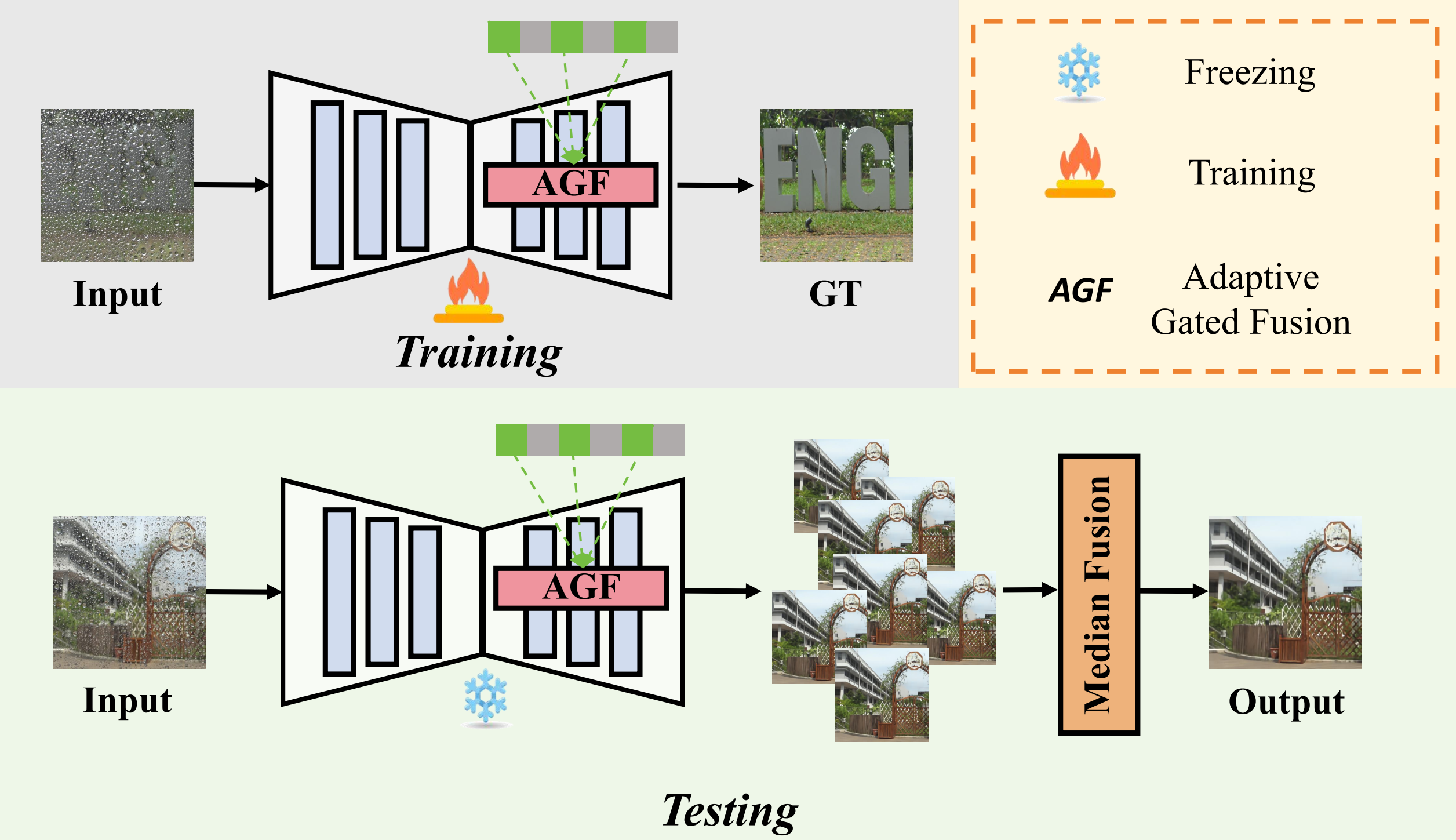}
    \caption{The pipeline of the method proposed by Team DGLTeam}
    \label{fig:DGLTeam}
\end{figure}

This team formulates multi-scene raindrop removal as a multi-task image restoration problem and builds their method upon a prompt-based all-in-one restoration framework~\cite{potlapalli2023promptir}, as illustrated in Fig.~\ref{fig:DGLTeam}. The network follows an encoder-decoder architecture augmented with a task prompt pool that stores task-relevant representations.

Given an input image, the model adaptively selects relevant prompts from the pool based on similarity and aggregates them into an instance-specific task representation. This representation is then integrated into the decoder via adaptive gated fusion, enabling the model to leverage both shared and task-specific information for restoration.

\noindent\textbf{Training Details.} 
The prompt‑driven model is trained on the official dataset using $128\times128$ patches with a batch size of~6.  Optimization employs the Adam algorithm (\(\beta_1=0.9,\beta_2=0.99\)) and an $\ell_1$ loss.  The learning rate is initialized at $4\times10^{-4}$ and decays to $1\times10^{-6}$ via a cosine schedule over roughly two million iterations (about 300~epochs).  Random flipping and rotation augmentations are applied.  Training runs for about three days on a single RTX~5090 GPU, and no external data are used.

\noindent\textbf{Testing Details.} 
Inference uses sliding‑window processing with $128\times128$ windows and 32‑pixel overlaps to mitigate border artifacts.  An eight‑fold geometric self‑ensemble (horizontal/vertical flips and $90^{\circ}$ rotations)~\cite{timofte2016seven} is performed, and the outputs are combined via median fusion rather than averaging.  A second median fusion across images from the same scene, followed by a weighted average with the input, further refines the result.

\noindent\textbf{Implementation Details.} 
The final model has about $42.6\,$M parameters and processes a $720\times480$ image in roughly $1.98\,$s on a single RTX~5090 GPU.  It does not utilize external datasets.

\subsection{MMAIrider}

\begin{figure}[t]
    \centering
    \includegraphics[width=1.0\linewidth]{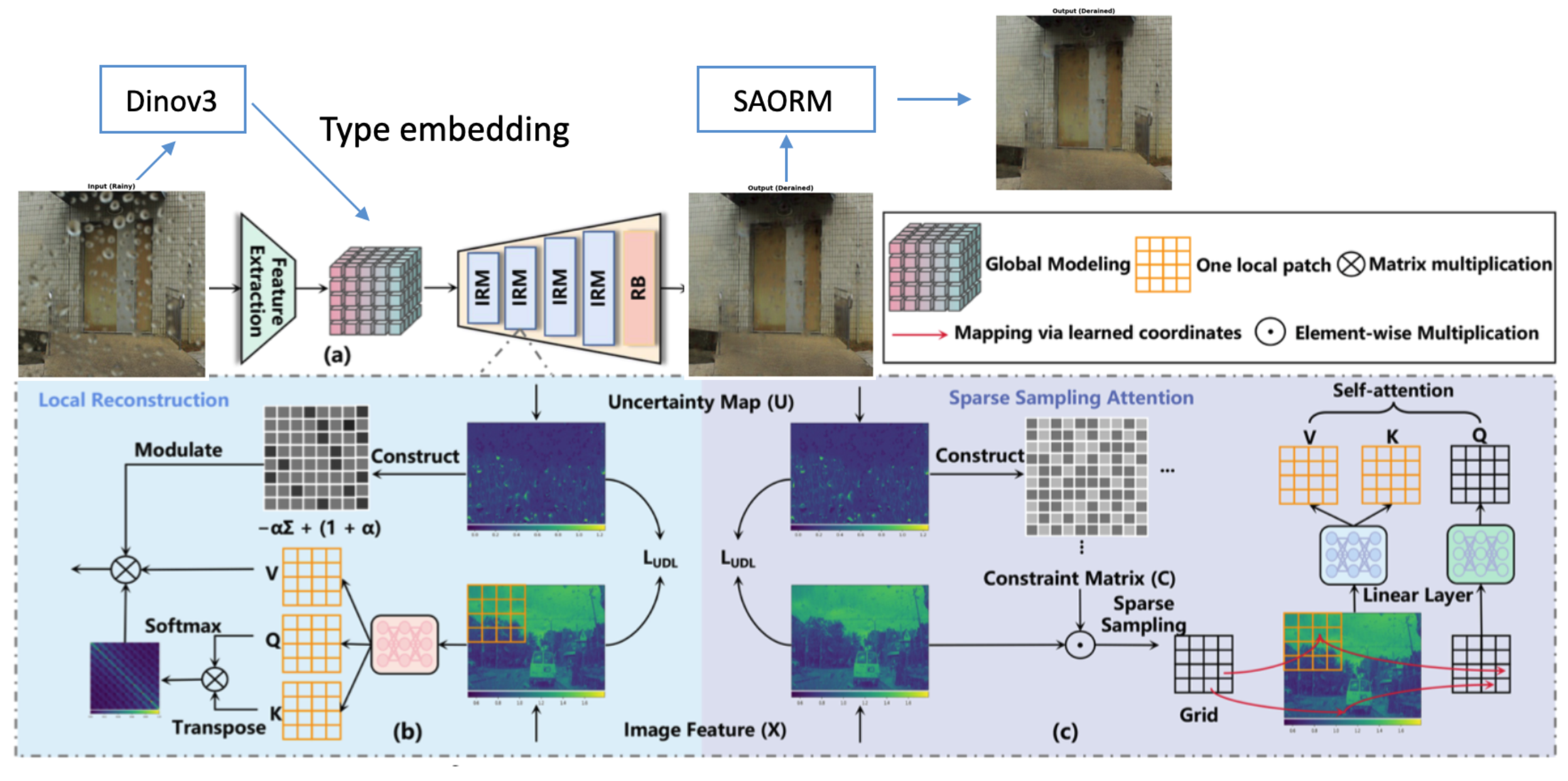}
    \caption{The pipeline of the method proposed by Team MMAIrider}
    \label{fig:MMAIrider}
\end{figure}

This team proposes a two-stage pipeline for dual-focused raindrop removal with uncertainty-guided restoration. The overall framework consists of a transformer-based deraining backbone followed by a refinement module, as illustrated in Fig.~\ref{fig:MMAIrider}.

In the first stage, a modified UDR-S2Former backbone is employed for initial raindrop removal. The network adopts a U-Net style transformer architecture with multi-scale encoder-decoder structure. A scene-guided modulation (SGM) mechanism is introduced at the bottleneck, where a learnable query bias is conditioned on scene type embeddings to adapt the attention behavior. In addition, deformable sampling attention and variance-guided local reconstruction are integrated into the decoder to enhance spatially adaptive restoration.

In the second stage, a Scene-Aware Omni-kernel Restoration~\cite{cui2024omni} Module (SAORM) is applied to refine the intermediate output. This module combines multiple branches, including local convolution, multi-scale dilated convolution, large-kernel convolution, and frequency-domain attention, to capture both local details and global contextual information. A dual-gating mechanism conditioned on scene embeddings further controls the fusion of restoration features.

Scene embeddings are obtained using a pretrained image encoder and are shared across both stages to guide scene-adaptive restoration.

\noindent\textbf{Training Details.} 
Training consists of a brief scene‑type embedding pretraining (12~epochs using DINOv3 features and a contrastive cosine loss) and a joint deraining and refinement stage.  The joint model is initialized from UDR‑S2Former weights and optimized for 300~epochs (about 24~hours) with a composite loss: $\ell_1$ + 0.3×MS‑SSIM + 0.3×auxiliary $\ell_1$ + 0.2×deep‑supervision $\ell_1$ + 0.5×uncertainty NLL.  AdamW with $\beta_1=0.9,\beta_2=0.999$ and weight decay~0.01 is used.  Layer‑wise learning rates are set with a base of $7\times10^{-4}$ for new modules and scaled down for earlier layers, following a cosine annealing schedule to $1\times10^{-7}$.  Training uses $256\times256$ crops, a batch size of~11, gradient clipping (max norm~1.0) and balanced sampling across four scene types.  All training is performed on a single RTX~5090 GPU with no extra data.

\noindent\textbf{Testing Details.} 
For inference, the image is divided into $256\times256$ tiles with 64‑pixel overlaps.  Four geometric test‑time augmentations (identity, horizontal flip, vertical flip and both flips) are applied.  After processing each tile under each augmentation, the inverse transforms are applied and the outputs are averaged to obtain the final restoration.

\noindent\textbf{Implementation Details.} 
The joint deraining and SAORM model has about $10.17\,$M trainable parameters.  Full‑resolution inference, including the 4‑fold TTA, takes roughly $1.15\,$s per image on an RTX~5090 GPU.  The code is implemented in Python with PyTorch 2.x and PyTorch Lightning, and no external data are used.

\subsection{Rain-SVNIT}

\begin{figure}[t]
    \centering
    \begin{subfigure}{0.49\linewidth}
        \centering
        \includegraphics[width=\linewidth]{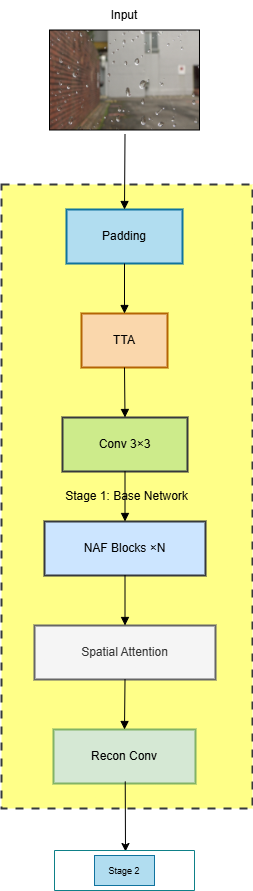}
        \caption{Stage 1: Global restoration}
        \label{fig:RainSVNIT_s1}
    \end{subfigure}
    \hfill
    \begin{subfigure}{0.49\linewidth}
        \centering
        \includegraphics[width=\linewidth]{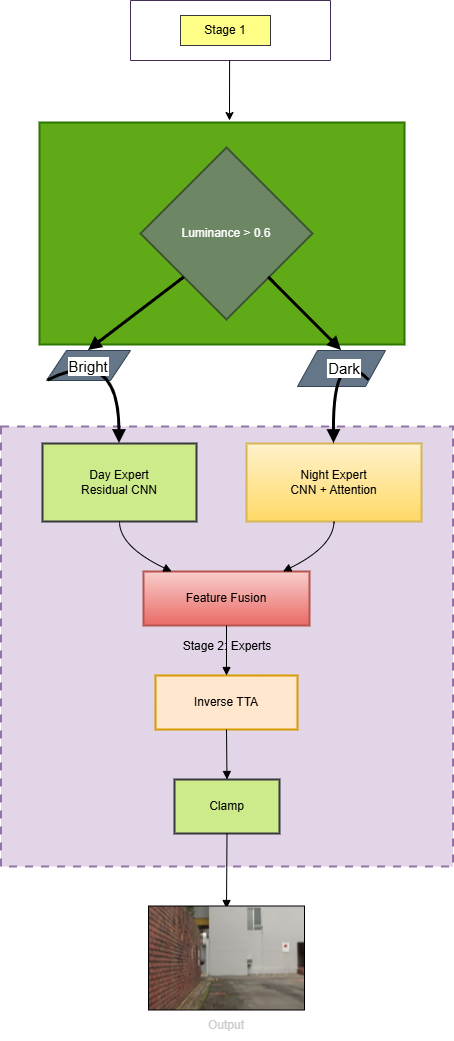}
        \caption{Stage 2: Luminance-based expert routing}
        \label{fig:RainSVNIT_s2}
    \end{subfigure}
    \caption{The pipeline of the method proposed by Team Rain-SVNIT. The framework consists of a Restormer-based global restoration stage followed by a luminance-based expert routing stage using specialized NAFNet models.}
    \label{fig:RainSVNIT}
\end{figure}

This team proposes a two-stage pipeline with luminance-based expert routing for dual-focused raindrop removal, as shown in Fig.~\ref{fig:RainSVNIT}. 

In the first stage (see Fig.~\ref{fig:RainSVNIT_s1}), a Restormer-based model~\cite{zamir2022restormer_deblur-transformer} performs global raindrop removal to produce an initial restoration. In the second stage (see Fig.~\ref{fig:RainSVNIT_s2}), the intermediate result is routed to specialized NAFNet experts~\cite{chen2022simpleNAFNet} according to image luminance, where separate models are designed for day and night scenarios. This routing mechanism enables domain-specific restoration and reduces cross-domain interference.

\noindent\textbf{Training Details.} 
The pipeline is trained in two separate stages.  Stage~1 fine‑tunes a Restormer model for 120~epochs (around 48~hours) using the Charbonnier loss on the official dataset.  Stage~2 trains two NAFNet experts for day and night scenes for 80~epochs (~24~hours) using a combination of Charbonnier and edge‑aware losses.  No additional data are used.

\noindent\textbf{Testing Details.} 
At inference, the luminance of the intermediate result is used to route the image to the appropriate expert.  Each expert prediction is further refined with an eight‑fold geometric self‑ensemble~\cite{timofte2016seven} comprising flips and rotations, and the ensemble results are averaged to produce the final output.

\noindent\textbf{Implementation Details.} 
The overall system contains roughly $26.1\,$M parameters and requires about $140\,$GFLOPs for a $256\times256$ patch.  With the 8× TTA, inference takes approximately $800\,$ms per image on an NVIDIA RTX~3090.  The models are trained in PyTorch using a cosine‑annealing learning rate and mixed‑precision training, and no external data are used.

\subsection{Just JiT}

\begin{figure}[t]
    \centering
    \includegraphics[width=1.0\linewidth]{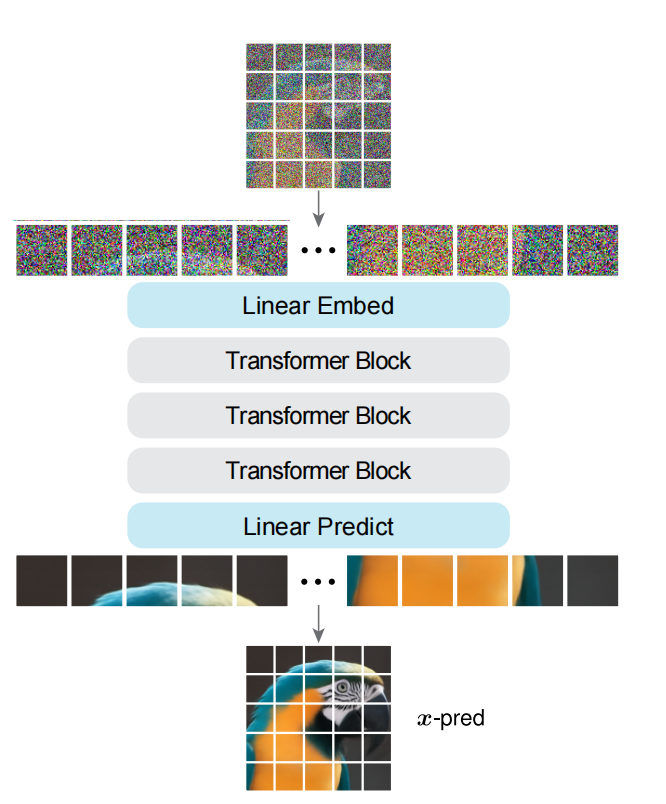}
    \caption{The pipeline of the method proposed by Team Just JiT}
    \label{fig:Just_JIT}
\end{figure}

This team proposes a scene-aware raindrop removal pipeline based on the Just Image Transformer (JiT)~\cite{li2025jit}, as illustrated in Fig.~\ref{fig:Just_JIT}. The method scales the backbone from JiT-B to JiT-H and introduces scene category information as conditional guidance for restoration. Specifically, the input images are divided into four scene categories, including day, night, background-focused, and raindrop-focused cases, and the corresponding scene labels are used to guide the transformer during restoration.

\noindent\textbf{Training Details.} 
Both JiT‑B and JiT‑H variants are trained on the official dual‑focused images for 600~epochs (about 48~hours) using $256\times256$ crops.  A ConvNeXt‑Tiny classifier~\cite{liu2022convnet} provides scene labels that condition the JiT transformer.  Optimization employs the AdamW optimizer with a base learning rate of $5\times10^{-5}$ and a cosine annealing schedule.  A dynamic hybrid loss combines Charbonnier, SSIM~\cite{SSIM} and LPIPS~\cite{zhang2018lpips} terms: the perceptual (LPIPS) weight is set to zero for the first 30\% of epochs and then linearly increased to its final value to balance fidelity and perceptual quality.  No extra data are used.

\noindent\textbf{Testing Details.} 
High‑resolution inference divides each image into overlapping $256\times256$ patches with a stride of 128~pixels.  A 2D Hanning window is multiplied with each restored patch before accumulation to suppress seam artifacts.  Outputs from JiT‑B and JiT‑H are averaged to obtain the final prediction.

\noindent\textbf{Implementation Details.} 
The JiT‑H model has approximately $953\,$M parameters and a computational cost of about $182\,$GFLOPs.  End‑to‑end inference with patch blending takes roughly $1.2\,$s per image on a single RTX~6000 GPU.  The code is implemented in PyTorch 2.0 using \texttt{torch.compile} and TF32 support, and no external data are used.

\subsection{PSU}

\begin{figure}[t]
    \centering
    \includegraphics[width=1.0\linewidth]{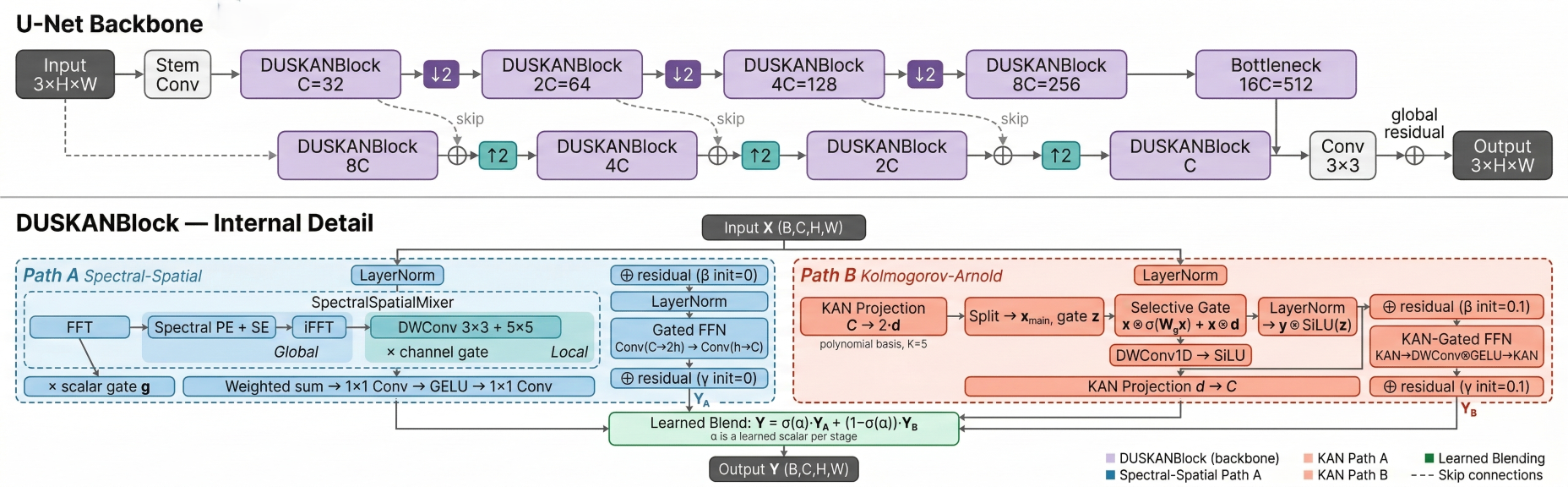}
    \caption{The pipeline of the method proposed by Team PSU}
    \label{fig:PSU}
\end{figure}

This team proposes DUSKAN, a dual-path neural architecture designed to capture both global frequency structures and local spatial details for dual-focused raindrop removal, as shown in Fig.~\ref{fig:PSU}. DUSKAN integrates a spectral-spatial branch, which processes features in both the spatial and frequency domains to handle global degradations like glare and rain scattering, and a Kolmogorov–Arnold branch, which uses adaptive polynomial activations to model complex non-linear relationships beyond standard fixed activations. The two branches operate in parallel within a U-Net backbone~\cite{unet} and are fused with learned gating weights, enabling the network to adaptively emphasize the most informative representation at each scale.

\noindent\textbf{Training Details.} 
The network is trained on $512\times512$ patches extracted from the official training data. Heavy augmentations, including random flips, rotations, and color jittering, are applied to improve robustness. The objective combines $\ell_1$ reconstruction, perceptual loss, and focal frequency loss to balance pixel-level fidelity and spectral consistency. Training runs for 500~epochs using the AdamW optimizer with a cosine learning rate schedule. No external datasets or pre-trained models are used.

\noindent\textbf{Testing Details.} 
Inference is performed on full-resolution images in a fully convolutional manner without sliding-window splitting. The model directly processes the entire image, leveraging its dual-branch design to recover both low-frequency global degradations and high-frequency details. No test-time augmentation or ensemble is applied.

\noindent\textbf{Implementation Details.} 
The DUSKAN model contains approximately $45.6\,$M parameters and requires about $493\,$GFLOPs per $512\times512$ input. Training is conducted using PyTorch on an NVIDIA GPU. The method uses only the provided competition dataset and does not incorporate any ensemble or external data.

\subsection{BITssvgg}

\begin{figure}[t]
    \centering
    \includegraphics[width=1.0\linewidth]{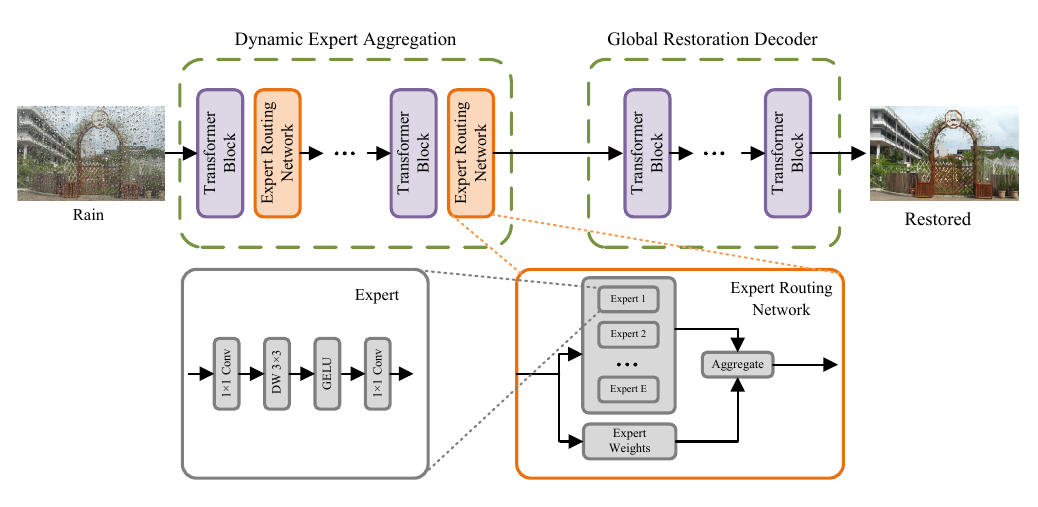}
    \caption{The pipeline of the method proposed by Team BITssvgg}
    \label{fig:BITssvgg}
\end{figure}

This team proposes DERNet, a dynamic expert routing network for raindrop removal, as illustrated in Fig.~\ref{fig:BITssvgg}. The model follows an encoder-decoder architecture that combines Transformer-based global modeling with adaptive local expert aggregation.

In the encoder, Transformer blocks are employed to capture long-range dependencies and global structural information. Meanwhile, dynamic expert routing modules are introduced to adaptively aggregate multiple lightweight convolutional experts for local restoration. Each expert consists of a compact depthwise convolutional structure, and the routing branch predicts expert-wise weights conditioned on the input features, enabling spatially adaptive processing for regions with different degradation levels.

The decoder adopts a Transformer-based design to progressively reconstruct the restored image. By integrating global contextual modeling with dynamic local expert selection, the network effectively handles spatially varying raindrop degradations.

\noindent\textbf{Training Details.} 
DERNet is trained for 180~epochs (about 26~hours) on the official dataset using patch‑based training and a plain $\ell_1$ reconstruction loss.  The Adam optimizer is used, and no extra data or ensemble methods are introduced.

\noindent\textbf{Testing Details.} 
Inference operates in a fully convolutional manner with appropriate padding to accommodate arbitrary image sizes, avoiding patch splitting.

\noindent\textbf{Implementation Details.} 
The network has approximately $16.87\,$M parameters and a computational cost of about $496\,$GFLOPs.  A $720\times480$ image is processed in roughly $0.4\,$s on two RTX~4090 GPUs.  Only the provided dataset is used.
\appendix

\subsection*{Organizers}
\label{organizers}
\noindent\textit{\textbf{Title:}} NTIRE 2026 The Second Challenge on Day and Night Raindrop Removal for Dual-Focused Images

\noindent\textit{\textbf{Members:}} \\
Xin Li\textsuperscript{1} (\textcolor{magenta}{xin.li@ustc.edu.cn}), \\
Yeying Jin\textsuperscript{2,3} (\textcolor{magenta}{jinyeying@u.nus.edu}),\\
Suhang Yao\textsuperscript{1} (\textcolor{magenta}{yshustc@mail.ustc.edu.cn}),\\
Beibei Lin\textsuperscript{2} (\textcolor{magenta}{beibei.lin@u.nus.edu}),\\
Zhaoxin Fan\textsuperscript{4} (\textcolor{magenta}{zhaoxinf@buaa.edu.cn}),\\
Wending Yan\textsuperscript{5} (\textcolor{magenta}{yan\_wending@swjtu.edu.cn}),\\
Xin Jin\textsuperscript{6} (\textcolor{magenta}{jinxin@eitech.edu.cn}), \\
Zongwei Wu\textsuperscript{7} (\textcolor{magenta}{zongwei.wu@uni-wuerzburg.de}), \\
Bingchen Li\textsuperscript{1} (\textcolor{magenta}{lbc31415926@mail.ustc.edu.cn}), \\
Peishu Shi\textsuperscript{2} (\textcolor{magenta}{peishushi222@gmail.com}),\\
Yufei Wang\textsuperscript{8} (\textcolor{magenta}{ywang25@snap.com}),\\
Yu Li\textsuperscript{9} (\textcolor{magenta}{liyu@idea.edu.cn}), \\
Zhibo Chen\textsuperscript{1} (\textcolor{magenta}{chenzhibo@ustc.edu.cn}), \\
Bihan Wen\textsuperscript{10} (\textcolor{magenta}{bihan.wen@ntu.edu.sg}), \\
Robby T. Tan\textsuperscript{2} (\textcolor{magenta}{robby.tan@nus.edu.sg}), \\
Radu Timofte\textsuperscript{7} (\textcolor{magenta}{Radu.Timofte@uni-wuerzburg.de})

\noindent\textit{\textbf{Affiliations:}}

\noindent\textsuperscript{1} University of Science and Technology of China

\noindent\textsuperscript{2} National University of Singapore

\noindent\textsuperscript{3} Tencent 

\noindent\textsuperscript{4} Beihang University

\noindent\textsuperscript{5} Southwest Jiaotong University

\noindent\textsuperscript{6} Eastern Institute of Technology, Ningbo

\noindent\textsuperscript{7} Computer Vision Lab, University of Würzburg

\noindent\textsuperscript{8} SparcAI Inc.

\noindent\textsuperscript{9} IDEA

\noindent\textsuperscript{10} Nanyang Technological University

% Team sections ordered according to the final ranking.

\subsection*{AIIA-Lab}

\noindent\textit{\textbf{Title:}}

\noindent\textit{\textbf{Members:}} Runzhe Li (\textcolor{magenta}{lirunzhe@stu.hit.edu.cn}), Kui Jiang, Zhaocheng Yu, Yiang Chen, Junjun Jiang, and Xianming Liu

\noindent\textit{\textbf{Affiliations:}}

\noindent Harbin Institute of Technology, China

\subsection*{raingod}

\noindent\textit{\textbf{Title:}}

\noindent\textit{\textbf{Members:}} Hongde Gu (\textcolor{magenta}{ghd@njust.edu.cn}), Zeliang Li, Mache You, Jiangxin Dong, Jinshan Pan

\noindent\textit{\textbf{Affiliations:}}

\noindent Nanjing University of Science and Technology

\subsection*{BUU\_CV}

\noindent\textit{\textbf{Title:}}

\noindent\textit{\textbf{Members:}} Qiyu Rong (\textcolor{magenta}{20231083510916@buu.edu.cn}), Bowen Shao, Hongyuan Jing, Mengmeng Zhang, Bo Ding, Hui Zhang, and Yi Ren

\noindent\textit{\textbf{Affiliations:}}

\noindent Beijing Union University

\subsection*{RetinexDualV2}

\noindent\textit{\textbf{Title: Physically-Grounded Dual Retinex for Generalized UHD Image Restoration}}

\noindent\textit{\textbf{Members:}} Mohab Kishawy (\textcolor{magenta}{kishawym@mcmaster.ca}), Jun Chen

\noindent\textit{\textbf{Affiliations:}}

\noindent Department of Electrical and Computer Engineering, McMaster University

\subsection*{ULR}

\noindent\textit{\textbf{Title:}}

\noindent\textit{\textbf{Members:}} Anh-Kiet Duong\textsuperscript{1} (\textcolor{magenta}{anh.duong@univ-lr.fr}), Petra Gomez-Krämer\textsuperscript{1}, and Jean-Michel Carozza\textsuperscript{2}

\noindent\textit{\textbf{Affiliations:}}

\noindent\textsuperscript{1}L3i Laboratory, La Rochelle University, France

\noindent\textsuperscript{2}LIENSs Laboratory, La Rochelle University, France

\subsection*{GU-day Mate}

\noindent\textit{\textbf{Title:}} 

\noindent\textit{\textbf{Members:}} Wangzhi Xing\textsuperscript{1} (\textcolor{magenta}{w.xing@griffith.edu.au}), Xin Lu\textsuperscript{2}, Enxuan Gu\textsuperscript{3}, Jingxi Zhang\textsuperscript{1}, Diqi Chen\textsuperscript{4}, Qiaosi Yi\textsuperscript{5}, and Bingcai Wei\textsuperscript{6}

\noindent\textit{\textbf{Affiliations:}}

\noindent\textsuperscript{1}Griffith University

\noindent\textsuperscript{2}SenseTime

\noindent\textsuperscript{3}Dalian University of Technology

\noindent\textsuperscript{4}Massey University

\noindent\textsuperscript{5}Hong Kong Polytechnic University

\noindent\textsuperscript{6}Wuhan University

\subsection*{Derain}

\noindent\textit{\textbf{Title:}} 

\noindent\textit{\textbf{Members:}} Wenjie Li (\textcolor{magenta}{cswjli@bupt.edu.cn}), Bowen Tie, Heng Guo, and Zhanyu Ma

\noindent\textit{\textbf{Affiliations:}}

\noindent Beijing University of Posts and Telecommunications

\subsection*{NTR}

\noindent\textit{\textbf{Title:}} 

\noindent\textit{\textbf{Members:}} Jiachen Tu (\textcolor{magenta}{jtu9@illinois.edu}), Guoyi Xu, Yaoxin Jiang, Cici Liu, and Yaokun Shi

\noindent\textit{\textbf{Affiliations:}}

\noindent University of Illinois at Urbana-Champaign, USA

\subsection*{Cidaut AI}

\noindent\textit{\textbf{Title:}} 

\noindent\textit{\textbf{Members:}} Paula Garrido Mellado (\textcolor{magenta}{paugar@cidaut.es}), Daniel Feijoo, Alvaro García Lara, and Marcos V. Conde

\noindent\textit{\textbf{Affiliations:}}

\noindent Cidaut AI

\subsection*{NCHU-CVLab}

\noindent\textit{\textbf{Title:}} 

\noindent\textit{\textbf{Members:}} Zhidong Zhu\textsuperscript{1} (\textcolor{magenta}{zhidong96@foxmail.com}), Bangshu Xiong\textsuperscript{2}, Qiaofeng Ou\textsuperscript{2}, Zhibo Rao\textsuperscript{2}, and Wei Li\textsuperscript{2}

\noindent\textit{\textbf{Affiliations:}}

\noindent\textsuperscript{1}Beihang University

\noindent\textsuperscript{2}Nanchang Hangkong University

\subsection*{NUDT-Deeplter}

\noindent\textit{\textbf{Title:}} 

\noindent\textit{\textbf{Members:}} Zida Zhang (\textcolor{magenta}{2475362322@qq.com}), Hui Geng, Qisheng Xu, Xuyao Deng, Changjian Wang, and Kele Xu

\noindent\textit{\textbf{Affiliations:}}

\noindent National University of Defense Technology

\subsection*{DGLTeam}

\noindent\textit{\textbf{Title:}} 

\noindent\textit{\textbf{Members:}} Guanglu Dong\textsuperscript{1} (\textcolor{magenta}{dongguanglu@stu.scu.edu.cn}), Qiyao Zhao\textsuperscript{2}, Tianheng Zheng\textsuperscript{1}, Chunlei Li\textsuperscript{3}, Lichao Mou\textsuperscript{3}, and Chao Ren\textsuperscript{1}

\noindent\textit{\textbf{Affiliations:}}

\noindent\textsuperscript{1}Sichuan University

\noindent\textsuperscript{2}Southwest University of Science and Technology

\noindent\textsuperscript{3}MedAI Technology (Wuxi) Co. Ltd.

\subsection*{MMAIrider}

\noindent\textit{\textbf{Title:}} 

\noindent\textit{\textbf{Members:}} Chang-De Peng, Chieh-Yu Tsai, Guan-Cheng Liu, and Li-Wei Kang (\textcolor{magenta}{lwkang@ntnu.edu.tw})

\noindent\textit{\textbf{Affiliations:}}

\noindent National Taiwan Normal University

\subsection*{Rain-SVNIT}

\noindent\textit{\textbf{Title:}} Dual-Focused Day-Night Raindrop Removal with Luminance-Based Expert Routing

\noindent\textit{\textbf{Members:}} Abhishek Rajak\textsuperscript{1} (\textcolor{magenta}{u24ec031eced@svnit.ac.in}), Milan Kumar Singh\textsuperscript{1}, Ankit Kumar\textsuperscript{1}, Dimple Sonone\textsuperscript{1}, Kishor Upla\textsuperscript{1}, and Kiran Raja\textsuperscript{2}

\noindent\textit{\textbf{Affiliations:}}

\noindent\textsuperscript{1}Sardar Vallabhbhai National Institute of Technology (SVNIT), Surat, India

\noindent\textsuperscript{2}Norwegian University of Science and Technology (NTNU), Gjøvik, Norway

\subsection*{Just JiT}

\noindent\textit{\textbf{Title:}} 

\noindent\textit{\textbf{Members:}} Huilin Zhao\textsuperscript{1} (\textcolor{magenta}{huilin.zhao@connect.polyu.hk}), Xing Xu\textsuperscript{2}, Chuan Chen\textsuperscript{3}, Yeming Lao\textsuperscript{1}, Wenjing Xun\textsuperscript{1}, and Li Yang\textsuperscript{4}

\noindent\textit{\textbf{Affiliations:}}

\noindent\textsuperscript{1}The Hong Kong Polytechnic University

\noindent\textsuperscript{2}Suzhou Institute for Advanced Research, University of Science and Technology of China

\noindent\textsuperscript{3}Technical University of Munich

\noindent\textsuperscript{4}Chizhou University

\subsection*{PSU}

\noindent\textit{\textbf{Title:}} DUSKAN: Dual Spectral Kolmogorov-Arnold Network

\noindent\textit{\textbf{Members:}} Bilel Benjdira (\textcolor{magenta}{bbenjdira@psu.edu.sa}), Anas M. Ali, Wadii Boulila

\noindent\textit{\textbf{Affiliations:}}

\noindent Robotics and Internet-of-Things Laboratory, Prince Sultan University, Riyadh 12435, Saudi Arabia

\subsection*{BITssvgg}

\noindent\textit{\textbf{Title:}} 

\noindent\textit{\textbf{Members:}} Hao Yang (\textcolor{magenta}{3120235187@bit.edu.cn}), Ruikun Zhang, and Liyuan Pan

\noindent\textit{\textbf{Affiliations:}}

\noindent Beijing Institute of Technology

\section*{Acknowledgments}
This work was partially supported by the Postdoctoral Fellowship Program of CPSF under Grant Number GZC20252293, the China Postdoctoral Science Foundation-Anhui Joint Support Program under Grant Number 2024T017AH, China Postdoctoral Science Foundation under Grant Number 2025M783529, Anhui Postdoctoral Scientific Research Program Foundation (No.2025A1015), the Fundamental Research Funds for the Central Universities (No. WK2100250064). This work was also partially supported by the Humboldt Foundation. We thank the NTIRE 2026 sponsors: OPPO, Kuaishou, and the University of Wurzburg (Computer Vision Lab).

{
    \small
    \bibliographystyle{ieeenat_fullname}
    \bibliography{main}

@String(CVPR= {IEEE Conf. Comput. Vis. Pattern Recog.})

@String(ICCV= {Int. Conf. Comput. Vis.})

@String(ECCV= {Eur. Conf. Comput. Vis.})

@String(ACMMM= {ACM Int. Conf. Multimedia})

@String(ICLR = {Int. Conf. Learn. Represent.})

@String(AAAI = {AAAI})

@String(CVPRW= {IEEE Conf. Comput. Vis. Pattern Recog. Worksh.})

@String(CVPR  = {CVPR})

@String(ICCV  = {ICCV})

@String(ECCV  = {ECCV})

@String(ACMMM = {ACM MM})

@String(ICLR  = {ICLR})

@String(CVPRW= {CVPRW})

@inproceedings{jin2024raindrop,
  title={Raindrop Clarity: A Dual-Focused Dataset for Day and Night Raindrop Removal},
  author={Jin, Yeying and Li, Xin and Wang, Jiadong and Zhang, Yan and Zhang, Malu},
  booktitle={European Conference on Computer Vision (ECCV)},
  pages={1--17},
  year={2024}
}

@article{liu2022convnet,
  title={A ConvNet for the 2020s},
  author={Liu, Zhuang and Mao, Hanzi and Wu, Chao-Yuan and Feichtenhofer, Christoph and Darrell, Trevor and Xie, Saining},
  journal={arXiv preprint arXiv:2201.03545},
  year={2022}
}

@article{li2025jit,
  title={Back to Basics: Let Denoising Generative Models Denoise},
  author={Li, Tianhong and He, Kaiming},
  journal={arXiv preprint arXiv:2511.13720},
  year={2025}
}

@inproceedings{chen2024rethinkingMSDT,
  title={Rethinking multi-scale representations in deep deraining transformer},
  author={Chen, Hongming and Chen, Xiang and Lu, Jiyang and Li, Yufeng},
  booktitle={AAAI},
  volume={38},
  number={2},
  pages={1046--1053},
  year={2024}
}

@inproceedings{sun2024restoringhistogram,
  title={Restoring images in adverse weather conditions via histogram transformer},
  author={Sun, Shangquan and Ren, Wenqi and Gao, Xinwei and Wang, Rui and Cao, Xiaochun},
  booktitle={ECCV},
  pages={111--129},
  year={2024},
  organization={Springer}
}

@inproceedings{chang2024uavrain1k,
  title={Uav-rain1k: A benchmark for raindrop removal from uav aerial imagery},
  author={Chang, Wenhui and Chen, Hongming and He, Xin and Chen, Xiang and Shen, Liangduo},
  booktitle={CVPR},
  pages={15--22},
  year={2024}
}

@inproceedings{rong2025STRRNet, 
title={{STRRNet}: Semantics-guided Two-stage Raindrop Removal 
  Network},
  author={Qiyu Rong and Hongyuan Jing and Mengmeng Zhang and Jinlong Li and Mengfei Han},
  booktitle={CVPRW},
  year={2025}
}

@inproceedings{lin2024nightrain,
  title={NightRain: Nighttime Video Deraining via Adaptive-Rain-Removal and Adaptive-Correction},
  author={Lin, Beibei and Jin, Yeying and Yan, Wending and Ye, Wei and Yuan, Yuan and Zhang, Shunli and Tan, Robby T},
  booktitle={Proceedings of the AAAI Conference on Artificial Intelligence},
  volume={38},
  number={4},
  pages={3378--3385},
  year={2024}
}

@inproceedings{jin2022unsupervised,
  title={Unsupervised night image enhancement: When layer decomposition meets light-effects suppression},
  author={Jin, Yeying and Yang, Wenhan and Tan, Robby T},
  booktitle={European Conference on Computer Vision},
  pages={404--421},
  year={2022},
  organization={Springer}
}

@inproceedings{jin2023enhancing,
  title={Enhancing visibility in nighttime haze images using guided apsf and gradient adaptive convolution},
  author={Jin, Yeying and Lin, Beibei and Yan, Wending and Yuan, Yuan and Ye, Wei and Tan, Robby T},
  booktitle={Proceedings of the 31st ACM International Conference on Multimedia},
  pages={2446--2457},
  year={2023}
}

@article{lin2024nighthaze,
  title={NightHaze: Nighttime Image Dehazing via Self-Prior Learning},
  author={Lin, Beibei and Jin, Yeying and Yan, Wending and Ye, Wei and Yuan, Yuan and Tan, Robby T},
  journal={arXiv preprint arXiv:2403.07408},
  year={2024}
}

@inproceedings{qian2018attentiveRaindropdataset,
  title={Attentive generative adversarial network for raindrop removal from a single image},
  author={Qian, Rui and Tan, Robby T and Yang, Wenhan and Su, Jiajun and Liu, Jiaying},
  booktitle={CVPR},
  pages={2482--2491},
  year={2018}
}

@inproceedings{porav2019canRaindropdataset,
  title={I can see clearly now: Image restoration via de-raining},
  author={Porav, Horia and Bruls, Tom and Newman, Paul},
  booktitle={ICRA},
  pages={7087--7093},
  year={2019},
  organization={IEEE}
}

@inproceedings{soboleva2021raindropsRaindropdataset,
  title={Raindrops on windshield: Dataset and lightweight gradient-based detection algorithm},
  author={Soboleva, Vera and Shipitko, Oleg},
  booktitle={IEEE SSCI},
  pages={1--7},
  year={2021},
  organization={IEEE}
}

@inproceedings{hao2019learningRaindropdataset,
  title={Learning from synthetic photorealistic raindrop for single image raindrop removal},
  author={Hao, Zhixiang and You, Shaodi and Li, Yu and Li, Kunming and Lu, Feng},
  booktitle={ICCV Workshops},
  pages={0--0},
  year={2019}
}

@inproceedings{chen2022simpleNAFNet,
  title={Simple baselines for image restoration},
  author={Chen, Liangyu and Chu, Xiaojie and Zhang, Xiangyu and Sun, Jian},
  booktitle={ECCV},
  pages={17--33},
  year={2022},
  organization={Springer}
}

@inproceedings{ntire26deepfake, 
title={{    Robust Deepfake Detection, NTIRE 2026 Challenge: Report    }}, 
author={    Hopf, Benedikt and  Timofte, Radu and others    },   
booktitle={Proceedings of the IEEE/CVF Conference on Computer Vision and Pattern Recognition (CVPR) Workshops},  
year = {2026} 
}

@inproceedings{ntire26hrdepth, 
title={{    NTIRE 2026 Challenge on High-Resolution Depth of non-Lambertian Surfaces    }}, 
author={    Zama Ramirez, Pierluigi and  Tosi, Fabio and  Di Stefano, Luigi and  Timofte, Radu and  Costanzino, Alex and  Poggi, Matteo and  Salti, Samuele and  Mattoccia, Stefano and others    },   
booktitle={Proceedings of the IEEE/CVF Conference on Computer Vision and Pattern Recognition (CVPR) Workshops},  
year = {2026} 
}

@inproceedings{ntire26raim_fusion, 
title={{    NTIRE 2026 The 3rd Restore Any Image Model (RAIM) Challenge: Multi-Exposure Image Fusion in Dynamic Scenes (Track2)    }}, 
author={    Qu, Lishen and  Liu, Yao and  Liang, Jie and  Zeng, Hui and  Dai, Wen and  Guan, Ya-nan and  Qin, Guanyi and  Zhou, Shihao and  Yang, Jufeng and  Zhang, Lei and  Timofte, Radu and others    },   
booktitle={Proceedings of the IEEE/CVF Conference on Computer Vision and Pattern Recognition (CVPR) Workshops},  
year = {2026} 
}

@inproceedings{ntire26raim_portrait, 
title={{    NTIRE 2026 The 3rd Restore Any Image Model (RAIM) Challenge: AI Flash Portrait (Track 3)    }}, 
author={    Guan, Ya-nan and  Zhang, Shaonan and  Guo, Hang and  Wang, Yawen and  Fan, Xinying and  Liang, Jie and  Zeng, Hui and  Qin, Guanyi and  Qu, Lishen and  Dai, Tao and  Xia, Shu-Tao and  Zhang, Lei and  Timofte, Radu and others    },   
booktitle={Proceedings of the IEEE/CVF Conference on Computer Vision and Pattern Recognition (CVPR) Workshops},  
year = {2026} 
}

@inproceedings{ntire26raim_piqa, 
title={{    NTIRE 2026 The 3rd Restore Any Image Model (RAIM) Challenge: Professional Image Quality Assessment (Track 1)    }}, 
author={    Qin, Guanyi and  Liang, Jie and  Zhang, Bingbing and  Qu, Lishen and  Guan, Ya-nan and  Zeng, Hui and  Zhang, Lei and  Timofte, Radu and others    },   
booktitle={Proceedings of the IEEE/CVF Conference on Computer Vision and Pattern Recognition (CVPR) Workshops},  
year = {2026} 
}

@inproceedings{ntire26lightsr, 
title={{    NTIRE 2026 Challenge on Light Field Image Super-Resolution: Methods and Results    }}, 
author={    Wang, Yingqian and  Liang, Zhengyu and  Zhang, Fengyuan and  Zhao, Wending and  Wang, Longguang and  Li, Juncheng and  Yang, Jungang and  Timofte, Radu and  Guo, Yulan and others    },   
booktitle={Proceedings of the IEEE/CVF Conference on Computer Vision and Pattern Recognition (CVPR) Workshops},  
year = {2026} 
}

@inproceedings{ntire263dsr, 
title={{    NTIRE 2026 Challenge on 3D Content Super-Resolution: Methods and Results    }}, 
author={    Wang, Longguang and  Guo, Yulan and  Wang, Yingqian and  Li, Juncheng and  Peng, Sida and  Zhang, Ye and  Timofte, Radu and  Chen, Minglin and  Wang, Yi and  Hu, Qibin and  Lei, Wenjie and others    },   
booktitle={Proceedings of the IEEE/CVF Conference on Computer Vision and Pattern Recognition (CVPR) Workshops},  
year = {2026} 
}

@inproceedings{ntire26videores, 
title={{    NTIRE 2026 Challenge on Bitstream-Corrupted Video Restoration: Methods and Results    }}, 
author={    Zou, Wenbin and  Liu, Tianyi and  Wu, Kejun and  Zhuang, Huiping and  Wu, Zongwei and  Zhou, Zhuyun and  Timofte, Radu and  others     },   booktitle={Proceedings of the IEEE/CVF Conference on Computer Vision and Pattern Recognition (CVPR) Workshops},  
year = {2026} 
}

@inproceedings{ntire26XAIGCqa, 
title={{    NTIRE 2026 X-AIGC Quality Assessment Challenge: Methods and Results    }}, 
author={    Liu, Xiaohong and  Min, Xiongkuo and  Zhai, Guangtao and  Hu, Qiang and  Cao, Jiezhang and  Zhou, Yu and  Sun, Wei and  Wen, Farong and  Xu, Zitong and  Zhou, Yingjie and  Duan, Huiyu and  Liu, Lu and  Wang, Jiarui and  Luo, Siqi and  Li, Chunyi and  Xu, Li and  Zhang, Zicheng and  Shi, Yue and  Wang, Yubo and  Zhang, Minghong and  Guo, Chunchao and  Hu, Zhichao and  Chen, Mingtao and  Wu, Xiele and  Ma, Xin and  Lv, Zhaohe and  Xue, Yuanhao and  Wang, Jiaqi and  Sha, Xinxing and  Timofte, Radu and  others    },   
booktitle={Proceedings of the IEEE/CVF Conference on Computer Vision and Pattern Recognition (CVPR) Workshops},  
year = {2026} 
}

@inproceedings{ntire26shadow, 
title={{    Advances in Single-Image Shadow Removal: Results from the NTIRE 2026 Challenge    }}, 
author={    Vasluianu, Florin-Alexandru and  Seizinger, Tim and  Zhou, Zhuyun and  Wu, Zongwei and  Timofte, Radu and  others     },   
booktitle={Proceedings of the IEEE/CVF Conference on Computer Vision and Pattern Recognition (CVPR) Workshops},  
year = {2026} 
}

@inproceedings{ntire26lightnorm, 
title={{    Learning-Based Ambient Lighting Normalization: NTIRE 2026 Challenge Results and Findings    }}, 
author={    Vasluianu, Florin-Alexandru and  Seizinger, Tim and  Chen, Jeffrey and  Zhou, Zhuyun and  Wu, Zongwei and  Timofte, Radu and  others    },   booktitle={Proceedings of the IEEE/CVF Conference on Computer Vision and Pattern Recognition (CVPR) Workshops},  
year = {2026} 
}

@inproceedings{ntire26bokeh, 
title={{    The First Controllable Bokeh Rendering Challenge at NTIRE 2026    }}, 
author={    Seizinger, Tim and  Vasluianu, Florin-Alexandru and  Conde, Marcos V. and  Chen, Jeffrey and  Zhou, Zhuyun and  Wu, Zongwei and  Timofte, Radu and  others    },   
booktitle={Proceedings of the IEEE/CVF Conference on Computer Vision and Pattern Recognition (CVPR) Workshops},  
year = {2026} 
}

@inproceedings{ntire26ripdetseg, 
title={{    NTIRE 2026 Rip Current Detection and Segmentation (RipDetSeg) Challenge Report    }}, 
author={    Dumitriu, Andrei and  Ralhan, Aakash and  Miron, Florin and  Tatui, Florin and  Ionescu, Radu Tudor and  Timofte, Radu and  others     },   booktitle={Proceedings of the IEEE/CVF Conference on Computer Vision and Pattern Recognition (CVPR) Workshops},  
year = {2026} 
}

@inproceedings{ntire26llie, 
title={{    Low Light Image Enhancement Challenge at NTIRE 2026    }}, 
author={    Ciubotariu, George and  S M A,  Sharif and  Rehman, Abdur and  Ali, Fayaz and  Naqvi, Rizwan Ali and  Conde, Marcos and  Timofte, Radu and others    },   
booktitle={Proceedings of the IEEE/CVF Conference on Computer Vision and Pattern Recognition (CVPR) Workshops},  
year = {2026} 
}

@inproceedings{ntire26highfps, 
title={{    High FPS Video Frame Interpolation Challenge at NTIRE 2026    }}, 
author={    Ciubotariu, George and  Zhou, Zhuyun and  Jin, Yeying and  Wu, Zongwei and  Timofte, Radu and  others    },   
booktitle={Proceedings of the IEEE/CVF Conference on Computer Vision and Pattern Recognition (CVPR) Workshops},  
year = {2026} 
}

@inproceedings{ntire26nthaze, 
title={{    NT-HAZE: A Benchmark Dataset for Realistic Night-time Image Dehazing    }}, 
author={    Ancuti, Radu and  Ancuti, Codruta and  Timofte, Radu and  Ancuti, Cosmin    },   
booktitle={Proceedings of the IEEE/CVF Conference on Computer Vision and Pattern Recognition (CVPR) Workshops},  
year = {2026} 
}

@inproceedings{li2025ntire,
  title={NTIRE 2025 challenge on day and night raindrop removal for dual-focused images: Methods and results},
  author={Li, Xin and Jin, Yeying and Jin, Xin and Wu, Zongwei and Li, Bingchen and Wang, Yufei and Yang, Wenhan and Li, Yu and Chen, Zhibo and Wen, Bihan and others},
  booktitle={Proceedings of the Computer Vision and Pattern Recognition Conference},
  pages={1172--1183},
  year={2025}
}

@article{xu2022codabench,
  title={Codabench: Flexible, easy-to-use, and reproducible meta-benchmark platform},
  author={Xu, Zhen and Escalera, Sergio and Pav{\~a}o, Adrien and Richard, Magali and Tu, Wei-Wei and Yao, Quanming and Zhao, Huan and Guyon, Isabelle},
  journal={Patterns},
  volume={3},
  number={7},
  year={2022},
  publisher={Elsevier}
}

@inproceedings{ntire26nthaze_rep, 
title={{    NTIRE 2026 Nighttime Image Dehazing Challenge Report    }}, 
author={    Ancuti, Radu and  Brateanu, Alexandru and  Vasluianu, Florin and  Balmez, Raul and  Orhei, Ciprian and  Ancuti, Codruta and  Timofte, Radu and  Ancuti, Cosmin and others    },   
booktitle={Proceedings of the IEEE/CVF Conference on Computer Vision and Pattern Recognition (CVPR) Workshops},  
year = {2026} 
}

@inproceedings{ntire26isp, 
title={{    NTIRE 2026 Challenge on Learned Smartphone ISP with Unpaired Data: Methods and Results    }}, 
author={    Perevozchikov, Georgy and  Vladimirov, Daniil and  Timofte, Radu and  others    },   
booktitle={Proceedings of the IEEE/CVF Conference on Computer Vision and Pattern Recognition (CVPR) Workshops},  
year = {2026} 
}

@inproceedings{ntire26ugcvideo, 
title={{    NTIRE 2026 Challenge on Short-form UGC Video Restoration in the Wild with Generative Models: Datasets, Methods and Results    }}, author={    Li, Xin and  Gong, Jiachao and  Wang, Xijun and  Xiong, Shiyao and  Li, Bingchen and  Yao, Suhang  and  Zhou, Chao and  Chen, Zhibo and  Timofte, Radu and others    },   
booktitle={Proceedings of the IEEE/CVF Conference on Computer Vision and Pattern Recognition (CVPR) Workshops},  
year = {2026} 
}

@inproceedings{ntire26srx4, 
title={{    The Fourth Challenge on Image Super-Resolution (×4) at NTIRE 2026: Benchmark Results and Method Overview    }}, 
author={    Chen, Zheng and  Liu, Kai and  Wang, Jingkai and  Yan, Xianglong and  Li, Jianze and  Zhang, Ziqing and  Gong, Jue and  Li, Jiatong and  Sun, Lei and  Liu, Xiaoyang and  Timofte, Radu and  Zhang, Yulun and others    },   
booktitle={Proceedings of the IEEE/CVF Conference on Computer Vision and Pattern Recognition (CVPR) Workshops},  
year = {2026} 
}

@inproceedings{ntire26retouching, 
title={{    Photography Retouching Transfer, NTIRE 2026 Challenge: Report    }}, 
author={    Elezabi, Omar and  V. Conde, Marcos and  Wu, Zongwei and  Jin, Yeying and  Timofte, Radu and others    },   
booktitle={Proceedings of the IEEE/CVF Conference on Computer Vision and Pattern Recognition (CVPR) Workshops},  
year = {2026} 
}

@inproceedings{ntire26rwsr, 
title={{    The First Challenge on Mobile Real-World Image Super-Resolution at NTIRE 2026: Benchmark Results and Method Overview    }}, 
author={    Li, Jiatong and  Chen, Zheng and  Liu, Kai and  Wang, Jingkai and  Zhou, Zihan and  Liu, Xiaoyang and  Zhu, Libo and  Timofte, Radu and  Zhang, Yulun and others    },   
booktitle={Proceedings of the IEEE/CVF Conference on Computer Vision and Pattern Recognition (CVPR) Workshops},  
year = {2026} 
}

@inproceedings{ntire26rsirsr, 
title={{    The First Challenge on Remote Sensing Infrared Image Super-Resolution at NTIRE 2026: Benchmark Results and Method Overview    }}, author={    Liu, Kai and  Yue, Haoyang and  Lin, Zeli and  Chen, Zheng and  Wang, Jingkai and  Gong, Jue and  Timofte, Radu and  Zhang, Yulun and  others    },   
booktitle={Proceedings of the IEEE/CVF Conference on Computer Vision and Pattern Recognition (CVPR) Workshops},  
year = {2026} 
}

@inproceedings{ntire26aigendet, 
title={{    NTIRE 2026 Challenge on Robust AI-Generated Image Detection in the Wild    }}, 
author={    Gushchin, Aleksandr and  Abud, Khaled and  Shumitskaya, Ekaterina and  Filippov, Artem and  Bychkov, Georgii and  Lavrushkin, Sergey and  Erofeev, Mikhail and  Antsiferova, Anastasia and  Chen, Changsheng and  Tan, Shunquan and  Timofte, Radu and  Vatolin, Dmitriy and others    },
booktitle={Proceedings of the IEEE/CVF Conference on Computer Vision and Pattern Recognition (CVPR) Workshops},  
year = {2026} 
}

@inproceedings{ntire26cdfsod, 
title={{    The Second Challenge on Cross-Domain Few-Shot Object Detection at NTIRE 2026: Methods and Results    }}, 
author={    Qiu, Xingyu and  Fu, Yuqian and  Geng, Jiawei and  Ren, Bin and  Pan, Jiancheng and  Wu, Zongwei and  Tang, Hao and  Fu, Yanwei and  Timofte, Radu and  Sebe, Nicu and  Elhoseiny, Mohamed and others    },   
booktitle={Proceedings of the IEEE/CVF Conference on Computer Vision and Pattern Recognition (CVPR) Workshops},  
year = {2026} 
}

@inproceedings{ntire26finrec, 
title={{    NTIRE 2026 Challenge on End-to-End Financial Receipt Restoration and Reasoning from Degraded Images: Datasets, Methods and Results    }}, author={    Guan, Bochen and  Li, Jinlong and  Yang, Kangning and  Ke, Chuang and  Cai, Jie and  Vasluianu, Florin and  Timofte, Radu and others    },   booktitle={Proceedings of the IEEE/CVF Conference on Computer Vision and Pattern Recognition (CVPR) Workshops},  
year = {2026} 
}

@inproceedings{ntire26faceres, 
title={{    The Second Challenge on Real-World Face Restoration at NTIRE 2026: Methods and Results    }}, 
author={    Wang, Jingkai and  Gong, Jue and  Chen, Zheng and  Liu, Kai and  Li, Jiatong and  Zhang, Yulun and  Timofte, Radu and  others    },
booktitle={Proceedings of the IEEE/CVF Conference on Computer Vision and Pattern Recognition (CVPR) Workshops},  
year = {2026} 
}

@inproceedings{ntire26reflection, 
title={{    NTIRE 2026 Challenge on Single Image Reflection Removal in the Wild: Datasets, Results, and Methods    }}, 
author={    Cai, Jie and  Yang, Kangning and  Li, Zhiyuan and  Vasluianu, Florin and  Timofte, Radu and others    },   
booktitle={Proceedings of the IEEE/CVF Conference on Computer Vision and Pattern Recognition (CVPR) Workshops},  
year = {2026} 
}

@inproceedings{ntire26anomalydet, 
title={{    NTIRE 2026  Challenge Report on Anomaly Detection of Face Enhancement for UGC Images    }}, 
author={    Zhong, Yan and   Ma,  Qiufang and  Wang, Zhen and  Jiang, Tingting and  Timofte, Radu and others    },   
booktitle={Proceedings of the IEEE/CVF Conference on Computer Vision and Pattern Recognition (CVPR) Workshops},  
year = {2026} 
}

@inproceedings{ntire26videosal, 
title={{    NTIRE 2026 Challenge on Video Saliency Prediction: Methods and Results    }}, 
author={    Moskalenko, Andrey and  Bryncev, Alexey and  Kosmynin, Ivan and  Shilovskaya, Kira and  Erofeev, Mikhail and  Vatolin, Dmitry and  Timofte, Radu and others    },   
booktitle={Proceedings of the IEEE/CVF Conference on Computer Vision and Pattern Recognition (CVPR) Workshops},  
year = {2026} 
}

@inproceedings{ntire26effsr, 
title={{    The Eleventh NTIRE 2026 Efficient Super-Resolution Challenge Report    }}, 
author={    Ren, Bin and  Guo, Hang and  Shu, Yan and  Ma, Jiaqi and  Cui, Ziteng and  Liu, Shuhong  and  Mei, Guofeng  and  Sun, Lei and  Wu, Zongwei and  Khan, Fahad Shahbaz and  Khan, Salman and  Timofte, Radu and  Li, Yawei and others    },   
booktitle={Proceedings of the IEEE/CVF Conference on Computer Vision and Pattern Recognition (CVPR) Workshops},  
year = {2026} 
}

@inproceedings{ntire26realx3d, 
title={{    3D Restoration and Reconstruction in Adverse Conditions: RealX3D Challenge Results    }}, 
author={    Liu, Shuhong and  Cui, Ziteng and  Bao, Chenyu and  Chu, Xuangeng and  Gu, Lin and  Ren, Bin and  Timofte, Radu and  Conde, Marcos V. and others    },   
booktitle={Proceedings of the IEEE/CVF Conference on Computer Vision and Pattern Recognition (CVPR) Workshops},  
year = {2026} 
}

@inproceedings{ntire26denoising, 
title={{    The Third Challenge on Image Denoising at NTIRE 2026: Methods and Results    }}, 
author={    Sun, Lei and  Guo, Hang and  Ren, Bin and  Su, Shaolin and  Wang, Xian and  Pani Paudel, Danda and  Van Gool, Luc and  Timofte, Radu and  Li, Yawei and others    },   
booktitle={Proceedings of the IEEE/CVF Conference on Computer Vision and Pattern Recognition (CVPR) Workshops},  
year = {2026} 
}

@inproceedings{ntire26aberration, 
title={{    NTIRE 2026 The First Challenge on Blind Computational Aberration Correction: Methods and Results    }}, 
author={    Sun, Lei and  Qian, Xiaolong and  Jiang, Qi and  Wang, Xian and  Gao, Yao and  Yang, Kailun and  Wang, Kaiwei and  Timofte, Radu and  Pani Paudel, Danda and  Van Gool, Luc and others    },   
booktitle={Proceedings of the IEEE/CVF Conference on Computer Vision and Pattern Recognition (CVPR) Workshops},  
year = {2026} 
}

@inproceedings{ntire26eventblurr, 
title={{    The Second Challenge on Event-Based Image Deblurring at NTIRE 2026: Methods and Results    }}, 
author={    Sun, Lei and  Li, Weilun and  Wang, Xian and  Li, Zhendong and  Shi, Letian and  Xu, Dannong and  Zhang, Deheng and  Hu, Mengshun and  Guo, Shuang and  Su, Shaolin and  Timofte, Radu and  Pani Paudel, Danda and  Van Gool, Luc and others    },   
booktitle={Proceedings of the IEEE/CVF Conference on Computer Vision and Pattern Recognition (CVPR) Workshops},  
year = {2026} 
}

@inproceedings{ntire26bursthdr, 
title={{    NTIRE 2026 Challenge on Efficient Burst HDR and Restoration: Datasets, Methods, and Results    }}, 
author={    Park, Hyunhee and  Park, Eunpil and  Lee, Sangmin and  Timofte, Radu and others    },   
booktitle={Proceedings of the IEEE/CVF Conference on Computer Vision and Pattern Recognition (CVPR) Workshops},  
year = {2026} 
}

@inproceedings{ntire26twilight, 
title={{    NTIRE 2026 Low-light Enhancement: Twilight Cowboy Challenge    }}, 
author={    Khalin, Aleksei and  Ershov, Egor and  Panshin, Artem and  Korchagin, Sergey and  Lobarev, Georgiy and  Terekhin, Arseniy and  Dorogova, Sofiia and  Shamsutdinov, Amir and  Mamedov, Yasin and  Khalfin, Bakhtiyar and  Sheludko, Bogdan and  Zilyaev, Emil and  Banić, Nikola and  Perevozchikov, Georgy and  Timofte, Radu and others    },   
booktitle={Proceedings of the IEEE/CVF Conference on Computer Vision and Pattern Recognition (CVPR) Workshops},  
year = {2026} 
}

@inproceedings{ntire26effllie, 
title={{    Efficient Low Light Image Enhancement: NTIRE 2026 Challenge Report    }}, 
author={    Yan, Jiebin  and  Tu, Chenyu  and  Lin, Qinghua and  WU, Zongwei and  Zhang , Weixia and  Wang, Zhihua and  Cao, Peibei and  Fang, Yuming  and  Liu, Xiaoning  and  Zhou, Zhuyun and  Timofte, Radu  and  others    },   
booktitle={Proceedings of the IEEE/CVF Conference on Computer Vision and Pattern Recognition (CVPR) Workshops},  
year = {2026} 
}

@article{potlapalli2023promptir,
  title={Promptir: Prompting for all-in-one image restoration},
  author={Potlapalli, Vaishnav and Zamir, Syed Waqas and Khan, Salman H and Shahbaz Khan, Fahad},
  journal={Advances in Neural Information Processing Systems},
  volume={36},
  pages={71275--71293},
  year={2023}
}

@inproceedings{valanarasu2022transweather,
  title={Transweather: Transformer-based restoration of images degraded by adverse weather conditions},
  author={Valanarasu, Jeya Maria Jose and Yasarla, Rajeev and Patel, Vishal M},
  booktitle={Proceedings of the IEEE/CVF conference on computer vision and pattern recognition},
  pages={2353--2363},
  year={2022}
}

@inproceedings{wu2024rainmamba,
  title={Rainmamba: Enhanced locality learning with state space models for video deraining},
  author={Wu, Hongtao and Yang, Yijun and Xu, Huihui and Wang, Weiming and Zhou, Jinni and Zhu, Lei},
  booktitle={ACMMM},
  pages={7881--7890},
  year={2024}
}

@inproceedings{zhang2018lpips,
  title={The unreasonable effectiveness of deep features as a perceptual metric},
  author={Zhang, Richard and Isola, Phillip and Efros, Alexei A and Shechtman, Eli and Wang, Oliver},
  booktitle={CVPR},
  pages={586--595},
  year={2018}
}

@inproceedings{unet,
  title={U-net: Convolutional networks for biomedical image segmentation},
  author={Ronneberger, Olaf and Fischer, Philipp and Brox, Thomas},
  booktitle={Medical image computing and computer-assisted intervention--MICCAI 2015: 18th international conference, Munich, Germany, October 5-9, 2015, proceedings, part III 18},
  pages={234--241},
  year={2015},
  organization={Springer}
}

@inproceedings{pang2020fan,
  title={Fan: Frequency aggregation network for real image super-resolution},
  author={Pang, Yingxue and Li, Xin and Jin, Xin and Wu, Yaojun and Liu, Jianzhao and Liu, Sen and Chen, Zhibo},
  booktitle={Computer Vision--ECCV 2020 Workshops: Glasgow, UK, August 23--28, 2020, Proceedings, Part III 16},
  pages={468--483},
  year={2020},
  organization={Springer}
}

@inproceedings{li2016rainRain12,
  title={Rain streak removal using layer priors},
  author={Li, Yu and Tan, Robby T and Guo, Xiaojie and Lu, Jiangbo and Brown, Michael S},
  booktitle={Proceedings of the IEEE conference on computer vision and pattern recognition},
  pages={2736--2744},
  year={2016}
}

@inproceedings{yang2017deepRain100HL,
  title={Deep joint rain detection and removal from a single image},
  author={Yang, Wenhan and Tan, Robby T and Feng, Jiashi and Liu, Jiaying and Guo, Zongming and Yan, Shuicheng},
  booktitle={Proceedings of the IEEE conference on computer vision and pattern recognition},
  pages={1357--1366},
  year={2017}
}

@inproceedings{fu2017removingDDN-data,
  title={Removing rain from single images via a deep detail network},
  author={Fu, Xueyang and Huang, Jiabin and Zeng, Delu and Huang, Yue and Ding, Xinghao and Paisley, John},
  booktitle={Proceedings of the IEEE conference on computer vision and pattern recognition},
  pages={3855--3863},
  year={2017}
}

@inproceedings{zhang2018densityDID-Data,
  title={Density-aware single image de-raining using a multi-stream dense network},
  author={Zhang, He and Patel, Vishal M},
  booktitle={Proceedings of the IEEE conference on computer vision and pattern recognition},
  pages={695--704},
  year={2018}
}

@inproceedings{qian2018attentiveRainDrop,
  title={Attentive generative adversarial network for raindrop removal from a single image},
  author={Qian, Rui and Tan, Robby T and Yang, Wenhan and Su, Jiajun and Liu, Jiaying},
  booktitle={Proceedings of the IEEE conference on computer vision and pattern recognition},
  pages={2482--2491},
  year={2018}
}

@inproceedings{li2019singleRain800,
  title={Single image deraining: A comprehensive benchmark analysis},
  author={Li, Siyuan and Araujo, Iago Breno and Ren, Wenqi and Wang, Zhangyang and Tokuda, Eric K and Junior, Roberto Hirata and Cesar-Junior, Roberto and Zhang, Jiawan and Guo, Xiaojie and Cao, Xiaochun},
  booktitle={Proceedings of the IEEE/CVF Conference on Computer Vision and Pattern Recognition},
  pages={3838--3847},
  year={2019}
}

@inproceedings{li2023learningDIL,
  title={Learning distortion invariant representation for image restoration from a causality perspective},
  author={Li, Xin and Li, Bingchen and Jin, Xin and Lan, Cuiling and Chen, Zhibo},
  booktitle={Proceedings of the IEEE/CVF Conference on Computer Vision and Pattern Recognition},
  pages={1714--1724},
  year={2023}
}

@inproceedings{wang2019spatialSPAData,
  title={Spatial attentive single-image deraining with a high quality real rain dataset},
  author={Wang, Tianyu and Yang, Xin and Xu, Ke and Chen, Shaozhe and Zhang, Qiang and Lau, Rynson WH},
  booktitle={Proceedings of the IEEE/CVF conference on computer vision and pattern recognition},
  pages={12270--12279},
  year={2019}
}

@inproceedings{hu2019depthRainCityScapes,
  title={Depth-attentional features for single-image rain removal},
  author={Hu, Xiaowei and Fu, Chi-Wing and Zhu, Lei and Heng, Pheng-Ann},
  booktitle={Proceedings of the IEEE/CVF Conference on computer vision and pattern recognition},
  pages={8022--8031},
  year={2019}
}

@inproceedings{li2019heavyOutdoor-Rain,
  title={Heavy rain image restoration: Integrating physics model and conditional adversarial learning},
  author={Li, Ruoteng and Cheong, Loong-Fah and Tan, Robby T},
  booktitle={Proceedings of the IEEE/CVF conference on computer vision and pattern recognition},
  pages={1633--1642},
  year={2019}
}

@inproceedings{jiang2020multiRain13K,
  title={Multi-scale progressive fusion network for single image deraining},
  author={Jiang, Kui and Wang, Zhongyuan and Yi, Peng and Chen, Chen and Huang, Baojin and Luo, Yimin and Ma, Jiayi and Jiang, Junjun},
  booktitle={Proceedings of the IEEE/CVF conference on computer vision and pattern recognition},
  pages={8346--8355},
  year={2020}
}

@inproceedings{yu2024sfiqa,
  title={Sf-iqa: Quality and similarity integration for ai generated image quality assessment},
  author={Yu, Zihao and Guan, Fengbin and Lu, Yiting and Li, Xin and Chen, Zhibo},
  booktitle={Proceedings of the IEEE/CVF Conference on Computer Vision and Pattern Recognition},
  pages={6692--6701},
  year={2024}
}

@inproceedings{quan2021removingRainDS,
  title={Removing raindrops and rain streaks in one go},
  author={Quan, Ruijie and Yu, Xin and Liang, Yuanzhi and Yang, Yi},
  booktitle={Proceedings of the IEEE/CVF conference on computer vision and pattern recognition},
  pages={9147--9156},
  year={2021}
}

@inproceedings{zou2024freqmambaDerain,
  title={Freqmamba: Viewing mamba from a frequency perspective for image deraining},
  author={Zou, Zhen and Yu, Hu and Huang, Jie and Zhao, Feng},
  booktitle={Proceedings of the 32nd ACM International Conference on Multimedia},
  pages={1905--1914},
  year={2024}
}

@inproceedings{lu2024aigcvqa,
  title={Aigc-vqa: A holistic perception metric for aigc video quality assessment},
  author={Lu, Yiting and Li, Xin and Li, Bingchen and Yu, Zihao and Guan, Fengbin and Wang, Xinrui and Liao, Ruling and Ye, Yan and Chen, Zhibo},
  booktitle={Proceedings of the IEEE/CVF Conference on Computer Vision and Pattern Recognition},
  pages={6384--6394},
  year={2024}
}

@inproceedings{tu2022maxim,
  title={Maxim: Multi-axis mlp for image processing},
  author={Tu, Zhengzhong and Talebi, Hossein and Zhang, Han and Yang, Feng and Milanfar, Peyman and Bovik, Alan and Li, Yinxiao},
  booktitle={Proceedings of the IEEE/CVF conference on computer vision and pattern recognition},
  pages={5769--5780},
  year={2022}
}

@inproceedings{li2020learningCNNderain,
  title={Learning disentangled feature representation for hybrid-distorted image restoration},
  author={Li, Xin and Jin, Xin and Lin, Jianxin and Liu, Sen and Wu, Yaojun and Yu, Tao and Zhou, Wei and Chen, Zhibo},
  booktitle={ECCV},
  pages={313--329},
  year={2020},
  organization={Springer}
}

@inproceedings{liang2021swinir,
  title={Swinir: Image restoration using swin transformer},
  author={Liang, Jingyun and Cao, Jiezhang and Sun, Guolei and Zhang, Kai and Van Gool, Luc and Timofte, Radu},
  booktitle={Proceedings of the IEEE/CVF international conference on computer vision},
  pages={1833--1844},
  year={2021}
}

@article{ozdenizci2023restoringweatherdiff,
  title={Restoring vision in adverse weather conditions with patch-based denoising diffusion models},
  author={{\"O}zdenizci, Ozan and Legenstein, Robert},
  journal={IEEE Transactions on Pattern Analysis and Machine Intelligence},
  volume={45},
  number={8},
  pages={10346--10357},
  year={2023},
  publisher={IEEE}
}

@article{shen2023rethinkingRainDiff,
  title={Rethinking real-world image deraining via an unpaired degradation-conditioned diffusion model},
  author={Shen, Yiyang and Wei, Mingqiang and Wang, Yongzhen and Fu, Xueyang and Qin, Jing},
  journal={arXiv preprint arXiv:2301.09430},
  year={2023}
}

@inproceedings{yang2017deep-rain100,
  title={Deep joint rain detection and removal from a single image},
  author={Yang, Wenhan and Tan, Robby T and Feng, Jiashi and Liu, Jiaying and Guo, Zongming and Yan, Shuicheng},
  booktitle={Proc. IEEE Conf. Comput. Vis. Pattern Recognit. workshops},
  pages={1357--1366},
  year={2017}
}

@article{SSIM,
  title={Image quality assessment: from error visibility to structural similarity},
  author={Wang, Zhou and Bovik, Alan C and Sheikh, Hamid R and Simoncelli, Eero P},
  journal={IEEE Trans. Image Process.},
  volume={13},
  number={4},
  pages={600--612},
  year={2004},
  publisher={IEEE}
}

@inproceedings{score,
  title={A kernelized Stein discrepancy for goodness-of-fit tests},
  author={Liu, Qiang and Lee, Jason and Jordan, Michael},
  booktitle={Int. Conf. Learn. Represent.},
  pages={276--284},
  year={2016},
  organization={PMLR}
}

@inproceedings{zamir2022restormer_deblur-transformer,
  title={Restormer: Efficient transformer for high-resolution image restoration},
  author={Zamir, Syed Waqas and Arora, Aditya and Khan, Salman and Hayat, Munawar and Khan, Fahad Shahbaz and Yang, Ming-Hsuan},
  booktitle={Proc. IEEE Conf. Comput. Vis. Pattern Recognit.},
  pages={5728--5739},
  year={2022}
}

@inproceedings{chen2024teachingt3diffusion,
  title={Teaching Tailored to Talent: Adverse Weather Restoration via Prompt Pool and Depth-Anything Constraint},
  author={Chen, Sixiang and Ye, Tian and Zhang, Kai and Xing, Zhaohu and Lin, Yunlong and Zhu, Lei},
  booktitle={European Conference on Computer Vision},
  pages={95--115},
  year={2024},
  organization={Springer}
}

@inproceedings{li2022hst,
  title={Hst: Hierarchical swin transformer for compressed image super-resolution},
  author={Li, Bingchen and Li, Xin and Lu, Yiting and Liu, Sen and Feng, Ruoyu and Chen, Zhibo},
  booktitle={European conference on computer vision},
  pages={651--668},
  year={2022},
  organization={Springer}
}

@inproceedings{cui2025adair,
  title={Adair: Adaptive all-in-one image restoration via frequency mining and modulation},
  author={Cui, Yuning and Zamir, Syed Waqas and Khan, Salman and Knoll, Alois and Shah, Mubarak and Khan, Fahad Shahbaz},
  booktitle={13th international conference on learning representations, ICLR 2025},
  pages={57335--57356},
  year={2025},
  organization={ICLR}
}

@article{tu2025d2ssft,
  title={Score-based self-supervised mri denoising},
  author={Tu, Jiachen and Shi, Yaokun and Lam, Fan},
  journal={arXiv preprint arXiv:2505.05631},
  year={2025}
}

@inproceedings{he2022mae,
  title={Masked autoencoders are scalable vision learners},
  author={He, Kaiming and Chen, Xinlei and Xie, Saining and Li, Yanghao and Doll{\'a}r, Piotr and Girshick, Ross},
  booktitle={Proceedings of the IEEE/CVF conference on computer vision and pattern recognition},
  pages={16000--16009},
  year={2022}
}

@article{tu2025scoremri,
  title={Score-based self-supervised mri denoising},
  author={Tu, Jiachen and Shi, Yaokun and Lam, Fan},
  journal={arXiv preprint arXiv:2505.05631},
  year={2025}
}

@inproceedings{cui2024omni,
  title={Omni-kernel network for image restoration},
  author={Cui, Yuning and Ren, Wenqi and Knoll, Alois},
  booktitle={Proceedings of the AAAI conference on artificial intelligence},
  volume={38},
  number={2},
  pages={1426--1434},
  year={2024}
}

@inproceedings{timofte2016seven,
  title={Seven ways to improve example-based single image super resolution},
  author={Timofte, Radu and Rothe, Rasmus and Van Gool, Luc},
  booktitle={Proceedings of the IEEE conference on computer vision and pattern recognition},
  pages={1865--1873},
  year={2016}
}

@misc{kishawy2026retinexdualv2physicallygroundeddualretinex,
      title={RetinexDualV2: Physically-Grounded Dual Retinex for Generalized UHD Image Restoration}, 
      author={Mohab Kishawy and Jun Chen},
      year={2026},
      eprint={2603.27979},
      archivePrefix={arXiv},
      primaryClass={cs.CV},
      url={https://arxiv.org/abs/2603.27979}, 
}

@misc{kishawy2025retinexdualretinexbaseddualnature,
      title={RetinexDual: Retinex-based Dual Nature Approach for Generalized Ultra-High-Definition Image Restoration}, 
      author={Mohab Kishawy and Ali Abdellatif Hussein and Jun Chen},
      year={2025},
      eprint={2508.04797},
      archivePrefix={arXiv},
      primaryClass={cs.CV},
      url={https://arxiv.org/abs/2508.04797}, 
}
}


\end{document}